\documentclass{article} 
\usepackage{iclr2023_conference,times}

\usepackage{hyperref}
\usepackage{url}

\usepackage{natbib}  
\usepackage{multirow}
\usepackage{graphicx}
\usepackage{caption}
\usepackage{url}
\usepackage{float}
\usepackage{array}
\usepackage{amsthm,amsmath,amssymb}
\usepackage{mathrsfs}
\usepackage{enumitem}
\usepackage{algorithm}
\usepackage{algorithmicx}
\usepackage{algpseudocode}
\usepackage{marvosym}
\usepackage{booktabs}
\usepackage{adjustbox}
\usepackage{epsfig}
\usepackage{tabularx}
\usepackage{colortbl}
\usepackage{pifont}
\usepackage{comment}
\usepackage{color}
\usepackage{verbatim}

\usepackage{amsmath,amsfonts,bm}

\def\eqref#1{equation~\ref{#1}}

\def\1{\bm{1}}

\DeclareMathAlphabet{\mathsfit}{\encodingdefault}{\sfdefault}{m}{sl}
\SetMathAlphabet{\mathsfit}{bold}{\encodingdefault}{\sfdefault}{bx}{n}

\title{Causal Inference via Nonlinear Variable Decorrelation for Healthcare Applications}

\author{Junda Wang \\
University of Massachusetts Amherst\\
Amherst, MA 01003, USA \\
\And
Weijian Li, Hanjia Lyu, Jiebo Luo\textsuperscript{*} \\
University of Rochester \\
Rochester, New York 14627, USA\\
\textsuperscript{*}\texttt{jluo@cs.rochester.edu}\\
\And
Han Wang \\
Shanghai Maritime University \\
Shanghai 201306, China\\
\And
Caroline Thirukumaran, Addisu Mesfin\\
University of Rochester Medical Center\\
Rochester, New York 14620, USA\\
}

\iclrfinalcopy
\begin{document}

\maketitle

\begin{abstract}
Causal inference and model interpretability research are gaining increasing attention, especially in the domains of healthcare and bioinformatics. Despite recent successes in this field, decorrelating features under nonlinear environments with human interpretable representations has not been adequately investigated. To address this issue, we introduce a novel method with a variable decorrelation regularizer to handle both linear and nonlinear confounding. Moreover, we employ association rules as new representations using association rule mining based on the original features to further proximate human decision patterns to increase model interpretability. Extensive experiments are conducted on four healthcare datasets (one synthetically generated and three real-world collections on different diseases). Quantitative results in comparison to baseline approaches on parameter estimation and causality computation indicate the model's superior performance. Furthermore, expert evaluation given by healthcare professionals validates the effectiveness and interpretability of the proposed model.

\end{abstract}

\section{Introduction}
With the rapid growth of Machine Learning (ML), healthcare ML research is becoming popular in the community. Such ML methods have shown encouraging capability for solving medically related problems, such as disease understanding, diagnosis, and treatment planning, by leveraging a large number of Electric Health Records (EHR). Although these methods bring benefits to both patients and healthcare professionals~\citep{herpertz2017challenge},  increasing concerns on judgment errors~\citep{royce2019teaching,gandhi2006missed} as well as deficiency of understanding the workflow of ML systems~\citep{croskerry2013mindless} have become major road-blockers for future development and deployment of ML-based healthcare systems. An important factor behind this difficulty is that the designed black-box ML models are often associated with a limited capacity for performance analysis~\citep{ahmad2018interpretable}. Therefore, building interpretable ML models for healthcare becomes an imperative research direction.

To improve the interpretability of black-box models, more and more methods to enhance model interpretability are emerging~\citep{du2019techniques, zafar2019dlime}. However, explanations of  black-box models often cannot be perfectly faithful to the original models and leave out much information which cannot be made sense of~\citep{rudin2019stop}. In addition, traditional ML models might be influenced by the data they are trained on.
In order to enhance the interpretability of the model and adapt it to human decision patterns, we introduce association rules in place of the original features.


Recently, most of the existing diagnostic algorithms focus on associative inference and are often not compatible with the situation caused by the incomplete distribution of datasets. Machine learning methods recognize diseases based on correlations and probability among patients' symptoms and medical history~\citep{zhang2021deep, kuang2020causal}, while doctors diagnose according to the best causal explanations corresponding to the symptoms~\citep{imbens2015causal}. Recently, several methods have been proposed to address the agnostic distribution, including domain generalization which is becoming one of the most prominent learning paradigms~\citep{muandet2013domain}. Another school of research examines the distribution shift issue from a causal perspective, such as causal transfer learning~\citep{rojas2018invariant} and Structural Causal Model (SCM)~\citep{pearl2009causal} to identify causal variables based on the conditional independence test. In spite of their advantageous analytical qualities, these approaches are rarely employed in high-dimensional real-world applications due to the complex causal graph and strict assumptions. More recently, some researchers focus on more general methods under the stability guarantee by variable decorrelation through sample reweighting~\citep{kuang2020stable, zhang2021deep, kuang2018stable, kuang2021balance}. They leveraged co-variate balancing to eliminate the impact of confounding, assessing the effect of the target feature by reweighting the data so that the distribution of covariates is equalized across different target feature values. However, their model are \textbf{limited to linear environments or binary datasets}.

In this paper, we attempt to address the aforementioned difficulties by developing a novel method that is inherently more interpretable and can be applied to \textbf{nonlinear environments} for stable prediction. We utilize an association rule mining algorithm to extract rules as model features, thereby enhancing our model's interpretability. To enable our model to operate in nonlinear environments, we model the relationships between features with a $F(x)$ function, and perform the Taylor expansion on the $F(x)$ function. The second norm of the parameters from the first derivative to the last derivative are considered as our regularizer. Experiments conducted on both synthetic and real-world datasets demonstrate the efficacy of our approach. Promising results in improving the estimation of model parameters, and the stability of prediction over varying distributions in a nonlinear environment demonstrate the superior performance of the proposed method to previous methods.

The main contributions of our work are as follows:

\begin{enumerate}
    \item[(1)] We expand the stable learning problem to a nonlinear environment under model misspecification and agnostic distribution so that stable learning can be widely applied in the real world;
    \item[(2)] We combine machine learning with association rules to help domain specialists understand the model and enhance the interpretability of the model; and
    \item[(3)] We demonstrate the superiority of our methods on synthetic and real-world datasets by calculating traditional metrics and causality. For medical datasets, we further invite specialized doctors to validate whether our model can produce the correct rules.
\end{enumerate}

\section{Related Work}
\subsection{Machine Learning Interpretability in Healthcare}

Increasing efforts have been devoted to Machine Learning (ML) interpretability research to facilitate ML research and development of real-world applications, especially in healthcare. Among them, Generalized Additive Models (GAM)~\citep{hastie2017generalized} are a set of classic methods with univariate terms providing straightforward interpretabilities. GA$^{2}$M-model~\citep{lou2013accurate} brings additional capability for real-world datasets with the selected interacting pairs based on GAMs. On the other hand, researchers focus on applying essentially interpretable models in healthcare domain. For example, \cite{lee2001review} apply association rules to extract knowledge as complementary information for physicians' diagnosis. They also provide some strategies for patients based on the interpretation of association rules. Lately, \cite{ahmed2021deep} apply association rules to detect major body organs in healthcare system. \cite{sornalakshmi2021efficient} reduce overhead communication when frequent data are extracted to improve association rules mining algorithm on healthcare datasets. However, the above models are still black-box models based on joint probability distribution without causal inference.

\subsection{Association Rule Mining}

Association rule mining is an important research direction that tries to identify interesting associations, frequent patterns, or causal structures~\citep{perccin2019arm,ordonez2006constraining}. In particular, association rules are able to discover predictive rules with numeric and categorical attributes. In diagnosis system, $X={x_{1}, x_{2},...,x_{n}}$ represents the set of all symptoms. An association rule, noted as $X \Rightarrow Y$, indicates the disease $Y$ is related to the symptoms $X$. Three metrics were proposed to evaluate the significance of rules: $\emph{support}(X)=P(X)$ is the probability that the set appears in the total item set; $\emph{confidence}(X \Rightarrow  Y)=\emph{support}(X\cup  Y)/\emph{support}(X)$ is a measure of reliability; $\emph{lift}(X \Rightarrow Y)=\emph{confidence}(X \Rightarrow Y)/\emph{support}(Y)$ reflects the correlation between $X$ and $Y$ in the association rules~\citep{bayardo1999mining}. The rules that satisfied the minimum support and confidence are called \textit{strong} association rules. Strong association rules, can also be divided into effective strong association rules and invalid strong association rules. How to extract strong association rules is an essential challenge. Apriori algorithm explores candidate-generation-and-test  to obtain strong association rules~\citep{borgelt2002induction}. Han proposed an effective method, the FP-Growth algorithm, to efficiently identity frequent patterns on large databases based on tree structures~\citep{han2000mining}. \cite{yuan2017improved} proposed an improved method based on the inherent defects of the Apriori algorithm by using a new mapping way and pruning frequent itemsets to improve efficiency. Association rules are interpretable models, whereas these methods always only consider extracting association rules based on connection instead of causality, and they do not combine rules and machine learning. In a diagnosis system, association rules are often inconsistent with the rules of doctors' diagnosis. Therefore, how to extract causal association rules that are consistent with doctors' diagnostic rules has become an important challenge.

\subsection{Causal Inference}

One key challenge in healthcare is the existence of both observed and unobserved confounders under different environments~\citep{cui2022stable}. Therefore, causal inference methods become popular for their natural fit to these problems. For example, causal inference methods with network and hierarchy structure allow researchers to ascribe causal explanations to data~\citep{pearl2018theoretical,pearl2009causal}. A completely constructed causal graph among various features based on an unconfoundedness assumption that helps to reduce the influence of confounders~\citep{ma2021assessing}. In addition, a Differentiated Variable Decorrelation (DVD) algorithm is proposed to eliminate the correlations of various variables in different environments~\citep{shen2020stable}. Moreover, \citet{xu2021stable} prove the effectiveness of stable learning and demonstrates the necessary of the stable prediction.





\noindent\textbf{Stable Learning} \quad Given various environments $\mathbf{e} \in E$ within datasets $D^{\mathbf{e}} = (X^{\mathbf{e}}, Y^{\mathbf{e}})$, the task is to train a predictive model under the environment $e_{i}$ which can achieve uniformly small error under the another environment $e_{j}$ by learning the causality between features $X^{e_{i}}$ and targets $Y^{e_{i}}$. Researchers propose the Deep Global Balancing Regression (DGBR) algorithm and  Decorrelated Weighting Regression (DWR) algorithm for stable prediction across unknown environments. They successively regard each variable as a treatment variable by using a balancing regularizer with theoretical guarantee~\citep{kuang2018stable, kuang2020stable, cui2022stable}.

In Equation~\ref{balancer}, $W$ is sample weight, $\mathbf{X}_{\cdot,j}$ is the $j^{th}$ variable in $\mathbf{X}$, and $\mathbf{X}_{\cdot,-j} = X/\{\mathbf{X}_{\cdot,j}\}$. With the global balancing regularizer in Equation~\ref{balancer}, a Global Balancing Regression algorithm is proposed to optimize global sample weights and causality for classification task.
\begin{equation}
\label{balancer}
\begin{array}{ll}
\min & \sum_{i=1}^{n} W_{i} \cdot \log \left(1+\exp \left(\left(1-2 Y_{i}\right) \cdot\left(\mathbf{X}_{i} \beta\right)\right)\right) \\
\text { s.t. } & \sum_{j=1}^{p}\left\|\frac{\mathbf{x}_{-j}^{T} \cdot(W \odot \mathbf{X}\cdot, j)}{W^{T} \cdot \mathbf{X}_{\cdot, j}}-\frac{\mathbf{X}_{i-j}^{T}(W \odot(1-\mathbf{X}\cdot, j))}{W^{T} \cdot\left(1-\mathbf{X}_{\cdot}, j\right)}\right\|_{2}^{2} \leq \lambda_{1}
\end{array}
\end{equation}

\begin{figure*}[t]
	\includegraphics[width=1\textwidth]{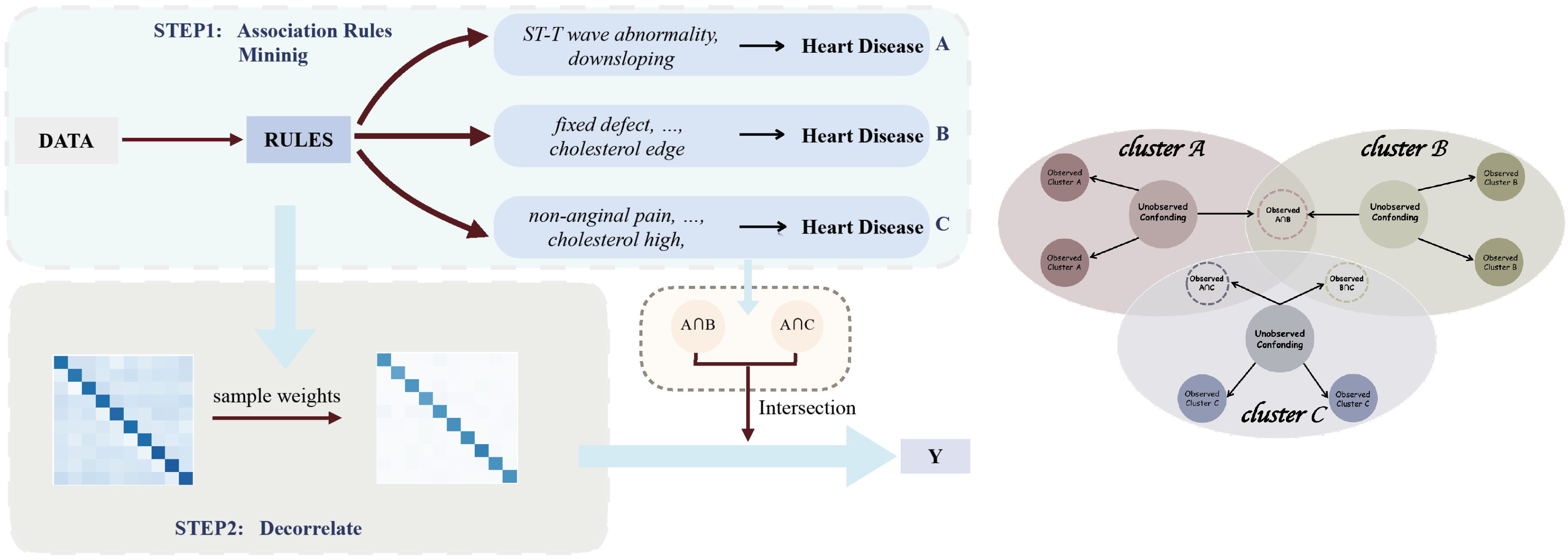}
	\centering
	\caption{Causality consists of intersecting regions and disjoint regions. Our model will extract rules from the original datasets and feed them into decorrelation parts to calculate the causality with mutual information.}
	\label{sample}
\end{figure*}

However, the above methods have some defects, making it difficult to deploy them on real world datasets. The regularizer of DGBR or DWR focuses on eliminating the linear confounding under various linear environments. In addition, such methods that forcibly delete mutual connections ignore information in the intersecting area. For example, as shown in Figure~\ref{sample}, forcibly eliminating the correlation may leave only three areas: $A, B, C$ and ignore the other areas. In fact, the causal effect of the three features should be the union of these areas, whereas DGBR may result in the loss of mutual information. 

\section{Method}

To achieve stable learning and  estimation with unbiased treatment effect, we make three assumptions for our model:

\begin{enumerate}[label=(\roman*)]
    \item The set of strong causality rules is a subset of the set of rules with strong correlations;
    \item There do not exist massive observed confounders that cause the diagnostic rules dependent on another unobserved confounding; and
    \item Causal rules tend to include smaller antecedents.
\end{enumerate}

The first and second assumptions imply that we can extract all the diagnostic rules with strong correlation, and then assign weights to these rules based on the causal effect. The third assumption implies that diagnosis rules with high confidence often result in overfitting due to redundant antecedents. Therefore, we prefer to choose more robust rules with shorter antecedents. In this paper, we propose an interpretable model based on association rules and causal inference for EHR datasets to obtain the causality between features through a three-stage process:

\textbf{A. Mining Association Rules and Transformation Rules:} First, we adopt the Apriori algorithm~\citep{agrawal1994fast} to obtain association rules and construct a rule matrix for positive and negative samples respectively to avoid asymmetrically distributed data. Rule representations $<\mathscr{A}_{i}, \mathscr{C}_{i}|\theta_{i}>$ are then constructed, where $\mathscr{A}_{i}$ is the antecedent of the rule $R_{i}$, $\mathscr{C}_{i}$ is the consequent of the rule and $\theta_{i}$ is the confidence of the rule, where
\begin{equation*}
    \{R\} = \{\cup_{i}R_{i}\}=\{\cup_{i}frequent(\mathscr{A}_{i} \cup \mathscr{C}_{i})\}
\end{equation*}
 According to the rules generated, rule sets $\cup_{i}\{X_{i}:<\mathscr{A}_{i}, \mathscr{C}_{i}|\theta_{i}>\}$ are built for dataset $D$ where each rule is considered as a feature. Rule sets are then transformed into one-zero matrix $X$ leveraging one-hot encoding.

\textbf{B. Selecting Rules:} Massive rules could be generated during the mining process, causing redundancy or even negative effects. To extract rules with strong correlations between features, we introduce an integer programming objective function:

\begin{equation}
\begin{array}{lll}
\label{selection}
\text { Min }& \left\|W\right\|_{2}^{2}+\left\|\max(0, 1- Yh(x)\right\|_{2}^{2}\\
&h(x) = (W^{T}X\odot rep(\mathbb{I}(R > 0), n) \theta + b) \\
\text { s.t. } & \sum_{i} \mathbb{I}(R_{i} > 0) \leq \lambda_{1} & \\
& \sum_{i} \mathbb{I}(R_{i} > 0) \geq \lambda_{2}\\
& \{R_{i}\} \in \text{\{0, 1\}} &
\end{array}\\
\end{equation}

where $\odot$ refers to the Hadamard product and $\mathbb{I}(R > 0)$ is the indicator function converting $R$, a set of rules, to a one-zero vector with $1*r$ dimension. The value of the indicator function equals to one when the frequency of the rule is more than zero, otherwise it equals to zero. $rep(\mathbb{I}(R > 0),n)$ function is defined to expand the vector $\mathbb{I}(R^{1*r} > 0)$ to a matrix with dimension $n*r$ where all rows are the same as the first row. $\lambda_{1}$ and $\lambda_{2}$ represent the bonds for the number of the selected rules. 

Since we only consider a binary-classification problem here, which is a common setting for most healthcare diagnosis problems, we take the inverse of the confidence of the negative class rule as the score. However, the number of rules mined by the association rule algorithm, e.g. Apriori, could be large, resulting an extremely high dimension of $R$ to be able to fit in Equation~\ref{selection}. Therefore, we propose to delete one redundant rule at a time during each n-fold cross-validation run based on a feature ranking criteria $w_{i}^{2}$. Details of rule selection can be found in Algorithm~\ref{rulereItemduce} in the Appendix under \emph{RulesSelection}.

Although redundant rules are removed accordingly, redundant items in rules could still impact the performance of the model. In addition, redundant items in different rules could be easily replaced with other rules. To solve this problem, we perform an iterative process to delete one item of each rule at a time which brings an updated $R$ with reduced dimension. Then we can reconstruct cross-validation sets and feed data into SVM models to get an average accuracy. The item that improves model's average accuracy the most will be deleted at every iteration. This step is summarized in Algorithm ~\ref{rulereItemduce} in the Appendix under \emph{ItemReduce}.

\textbf{C. Computing Causality Relationship:} To better handle real-world nonlinear relationships, we model nonlinear relationships under Taylor expansion with a function $\mathcal{F}(x)$ as is shown in Equation~\ref{taylor}. Each fixed point's derivatives can be considered as parameters to be solved by converting into a polynomial fitting problem due to the condition that two polynomials are equal only when both their degree and coefficients are the same. 

\begin{equation}
\begin{array}{ll}
\label{taylor}
x_{p_{1}} \sim & \mathcal{F}(x_{j})=f_{p_{1}p_{2}}(x_{p_{2}}(0))+f'_{p_{1}p_{2}}(x_{p_{2}}(0)) x_{p_{2}}+\\
& \frac{f''_{p_{1}p_{2}}(x_{p_{2}}(0))}{2!}x_{p_{2}}^{2}+\ldots+\frac{f^{(p)}_{p_{1}p_{2}}(x_{p_{2}}(0))}{p!}x_{p_{2}}^{p}+R_{p}(x_{p_{2}})
\end{array}
\end{equation}

The elimination of the impact of intersecting areas is achieved by balancing the weight $W$ as is represented in Equation~\ref{polynomial}. If $x_{p_{1}}$ and $x_{p_{2}}$ are independent and nonlinearly uncorrelated, the derivatives of their relation functions are all 0: $\left\|\{\mathscr{F}_{p_{2}\rightarrow p_{1}}\} \text { / } \{f_{p_{1}p_{2}}(x_{p_{2}}(0))\} \right\|$ = 0, and $\mathscr{F}_{p_{2}\rightarrow p_{1}}$ can be calculated using Equation~\ref{solution}

\begin{equation}
\begin{array}{ll}
\label{polynomial}
&\min_{\mathscr{F}_{p_{2}\rightarrow p_{1}}} R_{p}(x)^{2}\equiv \sum_{p_{1} \neq p_{2}} \sum_{i=1}^{n}\left[w_{i}x_{i p_{2}}-\mathcal{F}\left(w_{i}x_{i p_{1}}\right)\right]^{2} \\ \\
& \Rightarrow\mathscr{X}_{p_{2}}\left(w_{i}x_{p_{2}}\right) \mathscr{F}_{p_{2} \rightarrow p_{1}}=\mathscr{Y}_{p_{1}}
\end{array}
\end{equation}

\begin{equation*}
    \mathscr{X}_{p_{2}} = \left[\begin{array}{cccc}
n & \sum_{i=1}^{n} w_{i}x_{ip_{2}} &\cdots& \sum_{i=1}^{n} w_{i}^{k}x_{ip_{2}}^{k} \\
\sum_{i=1}^{n} w_{i}^{2}x_{ip_{2}} & \sum_{i=1}^{n} w_{i}^{2}x_{ip_{2}}^{2} &\cdots& \sum_{i=1}^{n} w_{i}^{k+1}x_{ip_{2}}^{k+1} \\
\vdots & \vdots &\ddots& \vdots \\
\sum_{i=1}^{n} w_{i}^{k}x_{ip_{2}}^{k} & \sum_{i=1}^{n} w_{i}^{k+1}x_{ip_{2}}^{k+1} &\cdots& \sum_{i=1}^{n} w_{i}^{2 k}x_{ip_{2}}^{2 k}
\end{array}\right]
\end{equation*}

\begin{equation*}
\mathscr{F}_{p_{2}\rightarrow p_{1}}=\left[\begin{array}{c}
f_{p_{1}p_{2}}(x_{p_{2}}(0)) \\
f'_{p_{1}p_{2}}(x_{p_{2}}(0)) \\
\vdots \\
f^{(p)}_{p_{1}p_{2}}(x_{p_{2}}(0))
\end{array}\right],
\mathscr{Y}_{p_{1}} = \left[\begin{array}{c}
\sum_{i=1}^{n} y_{i} \\
\sum_{j=1}^{n} x_{i} y_{i} \\
\vdots \\
\sum_{i=1}^{n} x_{i}^{k} y_{i}
\end{array}\right]
\end{equation*}

\begin{equation}
\label{solution}
\mathscr{F}_{p_{2}\rightarrow p_{1}} = \left(\mathscr{X}_{p_{2}}^{\mathrm{T}} \mathscr{X}_{p_{2}}\right)^{-1} \mathscr{X}_{p_{2}}^{\mathrm{T}} \mathscr{Y}_{p_{1}}
\end{equation}

Combined with Figure~\ref{sample}, physical meaning can be given to the above variables: $f(\mathbf{\theta})$ is the correlation between each feature and target; $C$ is the factor to expand the influence of the intersection area to get the real causality comparing with $W$ applied to eliminate the influence of the public area:

\begin{equation}
\begin{array}{llll}
\text{Min}&  \frac{1}{2} \beta^{T} \beta+ \sum_{i=1}^{n} (W_{i} + C)\max \left(0,1-y_{i}\left(\beta^{T} \phi\left(x_{i}\right)+b\right)\right) \\

&\|\mathscr{F}_{p_{2}\rightarrow p_{1}, i>0}^{(i)}\|_{2}^{2} \leq \gamma,
\|W\|_{2}^{2} \leq \lambda_{5},\left(\sum_{k=1}^{n} W_{k}-1\right)^{2} \leq \lambda_{6}&
\end{array}
\end{equation}

When we have a smaller $\gamma$ value, the difference between $\beta$ and the true correlation coefficient (disjoint region and the target) will become smaller, resulting greater mutual information loss.

\section{Experiment}

\subsection{Validation on A Synthetic Dataset}

To examine the proposed constraints' effect on eliminating linear and nonlinear connotation relationships, we follow previous work~\citep{kuang2020stable} to conduct evaluations on synthetically generated datasets. The details of experiment settings and baseline methods can be found in the Appendix (~\ref{exp_datasets}). Notice that a different objective function~\ref{regression} is built for regression task, where $W_{i}$ is the sample weight. In this experiment, we only expand two terms by the Taylor expansion.

\begin{equation}
\label{regression}
\begin{array}{r}
\min _{w, b, \zeta, \zeta^*} \frac{1}{2} w^T w+ \sum_{i=1}^n (C+W_{i})\left(\zeta_i+\zeta_i^*\right) \\
\text { subject to } y_i-w^T \phi\left(x_i\right)-b \leq \varepsilon+\zeta_i \\
w^T \phi\left(x_i\right)+b-y_i \leq \varepsilon+\zeta_i^* \\
\zeta_i, \zeta_i^* \geq 0, i=1, \ldots, n\\
\|\mathscr{F}_{p_{2}\rightarrow p_{1}, i>0}^{(i)}\|_{2}^{2} \leq \gamma\\
\|W\|_{2}^{2} \leq \lambda_{1},\left(\sum_{k=1}^{n} W_{k}-1\right)^{2} \leq \lambda_{2}
\end{array}
\end{equation}

\begin{figure*}
\begin{minipage}[t]{0.33\linewidth}
    \centering
    \includegraphics[scale=0.33]{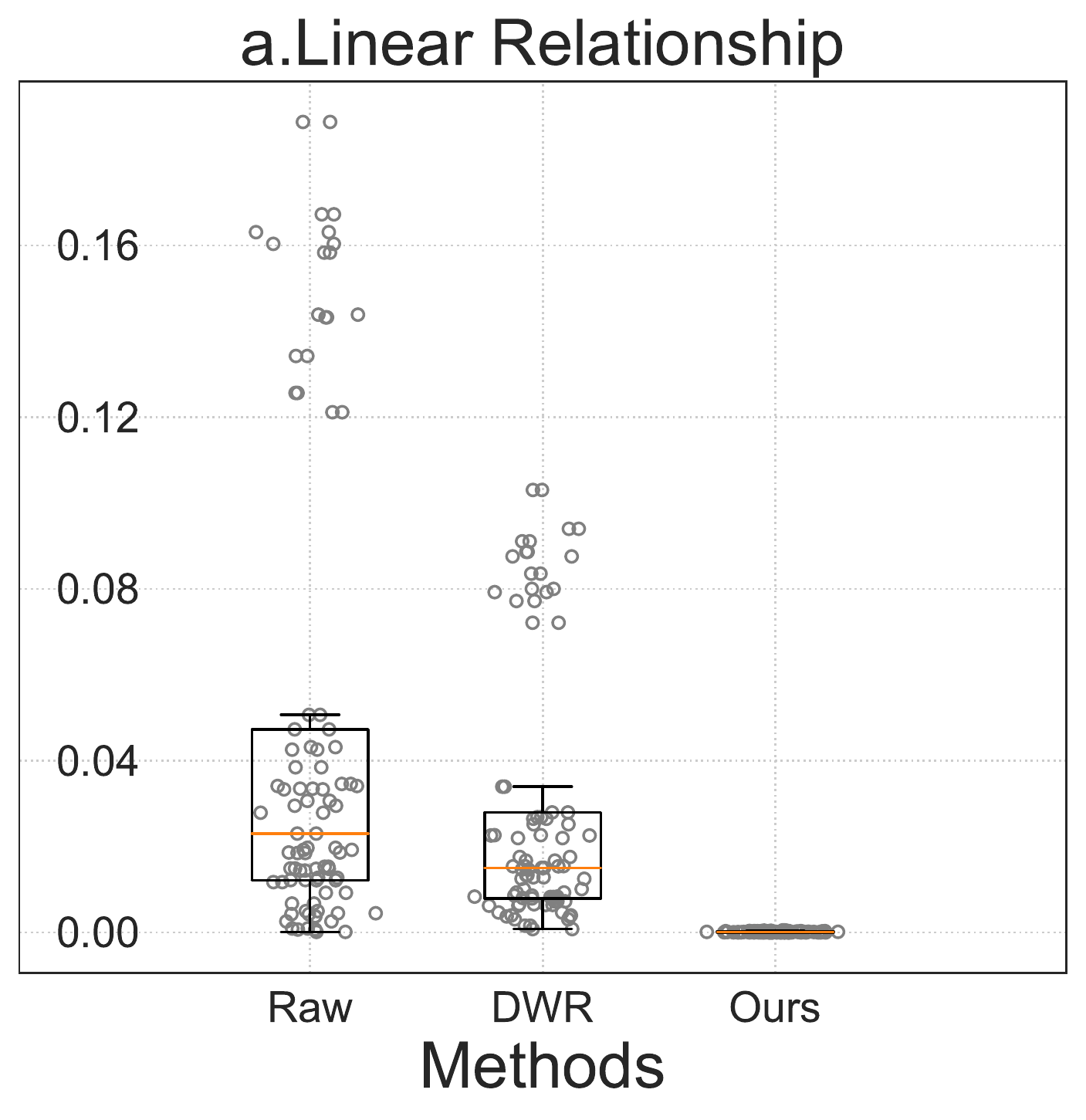}
  \end{minipage}
  \begin{minipage}[t]{0.33\linewidth}
    \centering
    \includegraphics[scale=0.33]{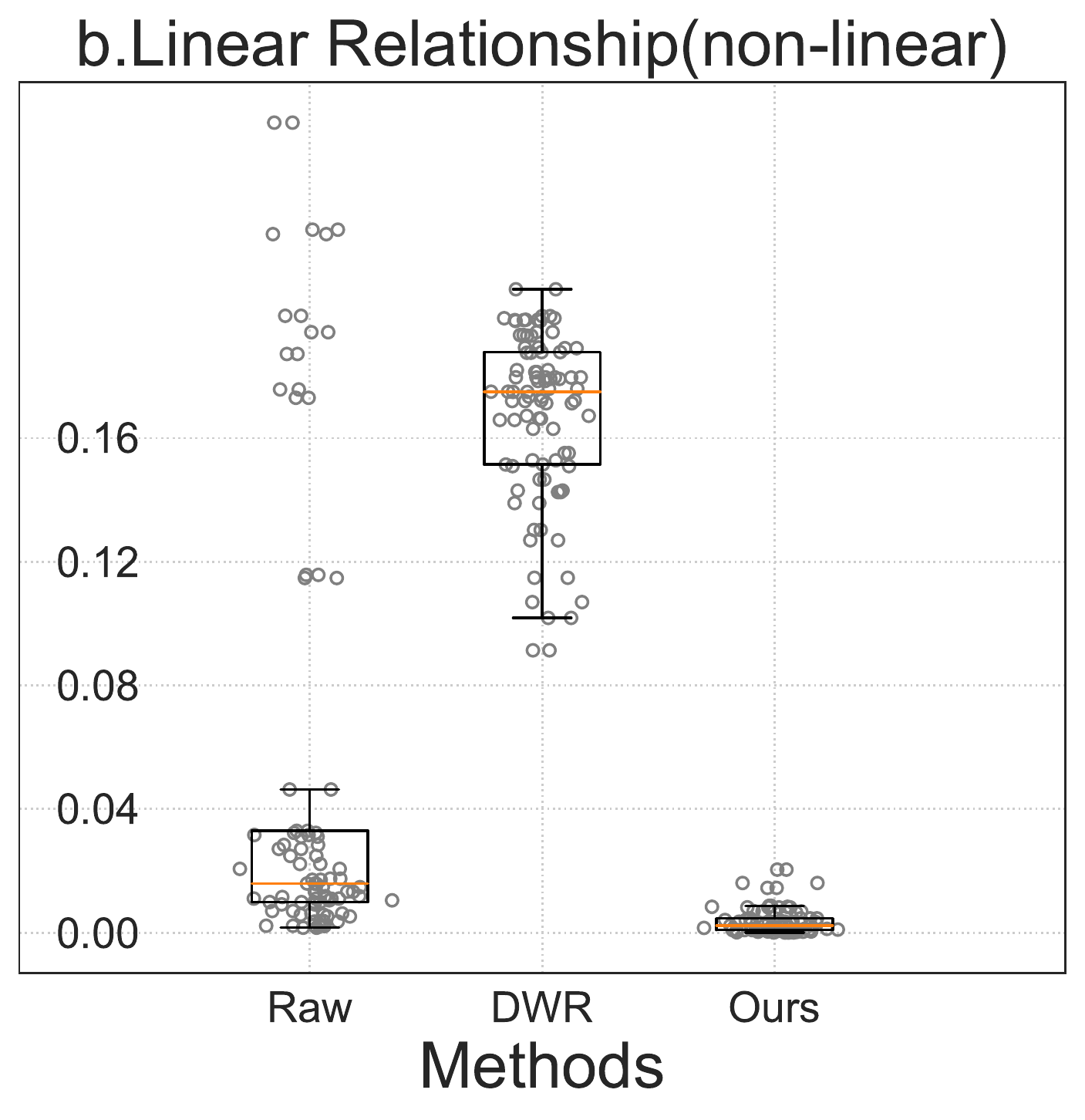}
  \end{minipage}
  \begin{minipage}[t]{0.33\linewidth}
    \centering
    \includegraphics[scale=0.33]{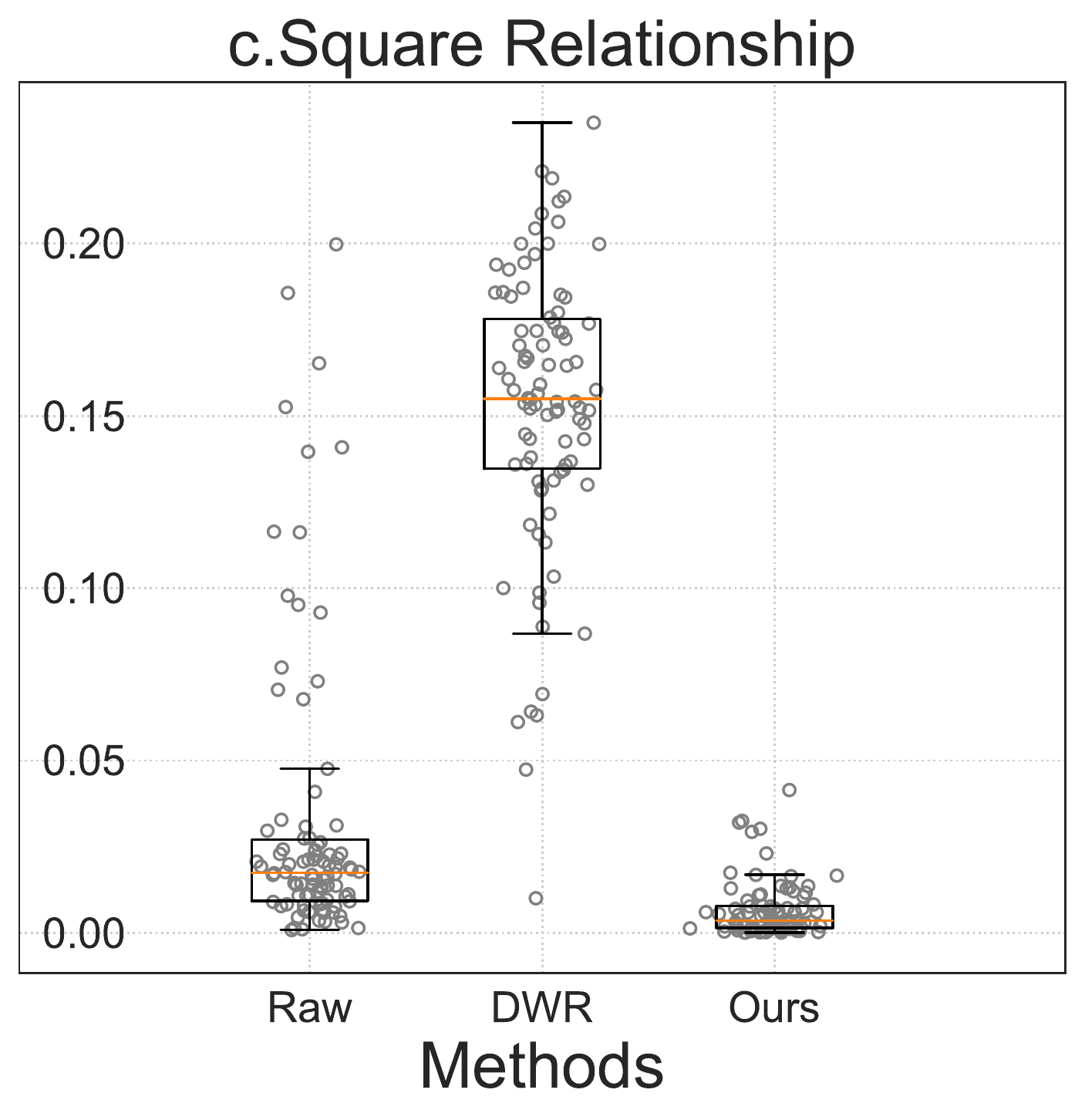}
  \end{minipage}
  \begin{minipage}[t]{0.33\linewidth}
    \centering
    \includegraphics[scale=0.33]{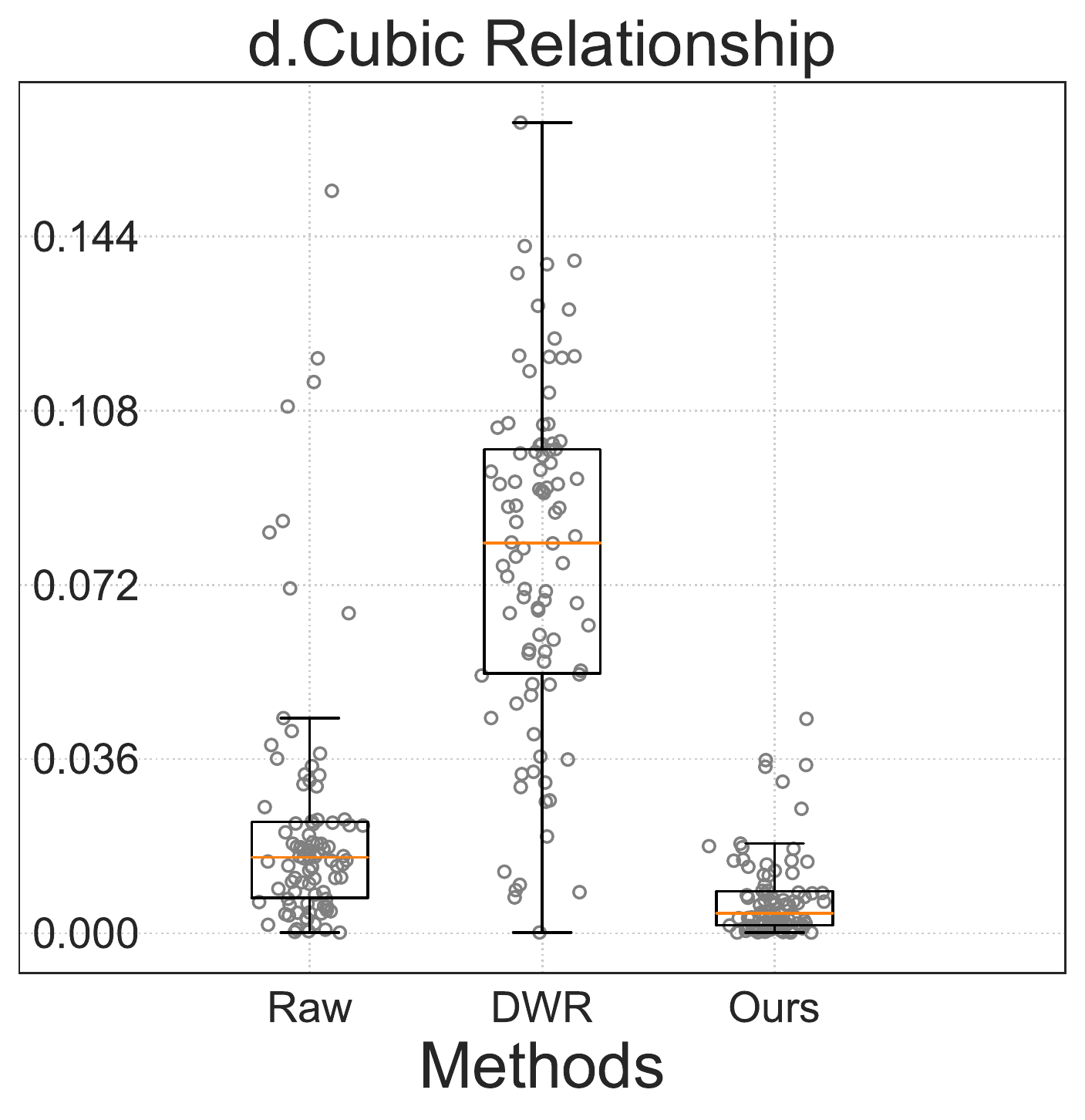}
  \end{minipage}%
  \begin{minipage}[t]{0.33\linewidth}
    \centering
    \includegraphics[scale=0.33]{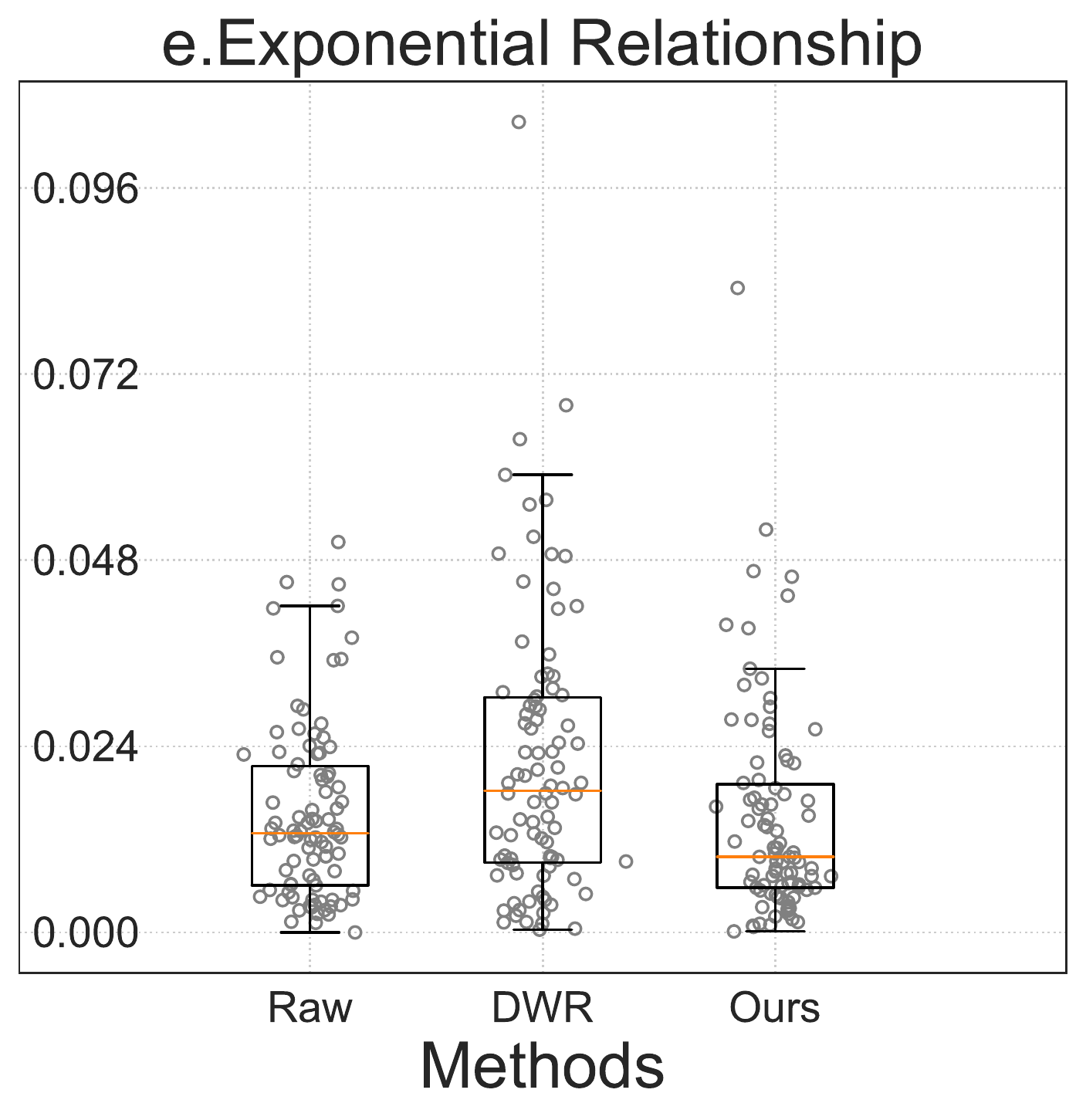}
  \end{minipage}
  \begin{minipage}[t]{0.33\linewidth}
    \centering
    \includegraphics[scale=0.33]{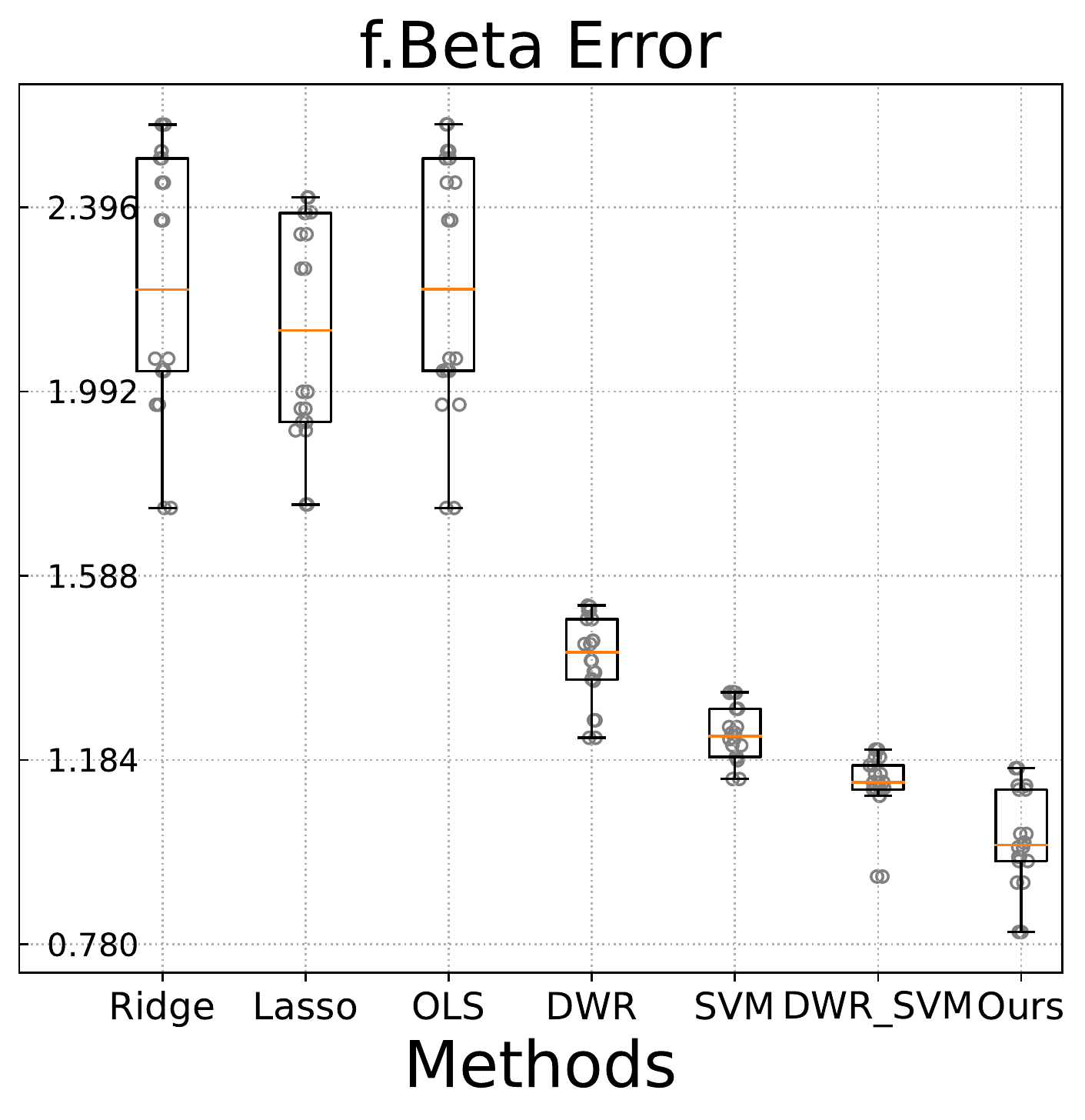}
  \end{minipage}%
  \caption{Figures (a)-(d) describe the distribution of the Pearson Coefficient values among various relationships. Figure (a) reports the $\beta$ errors of different models. Figure (f) is under a linear environment and other figures are under nonlinear environments. Our model is able to provide the greatest reduction of both linear and nonlinear relationships. }
  \vspace{-2mm}
  \label{beta_RMS}
\end{figure*}

\subsubsection{Results}
To compare two kinds of regularizers, we apply Pearson Correlation to calculate the relationship strength among features. Since Pearson Correlation can only describe linear relationship, we construct nonlinear pairs $\mathbf{WV_{i}}$ with $\mathbf{(WV)^{2}_{j}}$, $\mathbf{(WV)^{3}_{j}}$ and $exp(\mathbf{WV}_{j})$ in addition to $\mathbf{WV_{i}}$ with $\mathbf{WV_{j}}$. The result can be found in Figure~\ref{beta_RMS}. Both DWR and the proposed regularizer can handle pure linear relationships (experimental environment(A)) but some improvement is achieved from the proposed regularizer. As we add nonlinear relationships to the linear experimental environment, DWR start to have difficulty with the linear relationship part while the proposed method is still able to reduce a large amount of the relationships. For nonlinear environments, compared with the original unweighted dataset, DWR unexpectedly increases nonlinear relationships where there are no existing nonlinear relationships (square, cubic and exponential). Instead, our model can deal with nonlinear relationships and reduce nonlinear relationships. 

We conduct a series of ablative studies to evaluate the stability of our model. Table~\ref{Ablative} summarizes the results of the experiment with various values of $C$ and Lagrange penalty operators $\gamma$, and $\lambda$. For each sell, we fix the Lagrange penalty operators and increase $C$ to calculate the $\beta$ errors and RMSE errors. The higher $C$ indicates higher integrated mutual information is fed into the model and magnifies the impact of confounding. Higher $\gamma$ values will reduce more confounding effects and diminish mutual information. Here we choose the best parameters ($\gamma=600, \lambda=0.0005, C=0.5$) based on the smallest RMSE  also with a smaller $\beta$ error comparing to ($\gamma=1000, \lambda=0.0005, C=0.5$) which has the same RMSE.

To further confirm that the coefficients estimated by our model are based on causality, we repeat experiments 50 times to calculate $\sum{\|\beta -\hat{\beta}\|}$, where $\beta$ and $\hat{\beta}$ represent the true value and estimated parameters, respectively. In Figure~\ref{beta_RMS}, we find that the difference between the estimated parameters and the true values is smaller with our model, compared to other models in the nonlinear environment.
Notice that our model achieves much smaller distribution variance as well as much smaller average values of $\beta$ errors comparing to baselines. Although the regularizer of DWR can solve the stable problem in linear environments, it retains or expands nonlinear confounding in the nonlinear environments. From the above results, we find that our model is able to reduce correlations among all predictors and avoid being affected by nonlinear confounding, resulting a reduced estimation bias in more general environments.

\begin{table*}[!t]
    \centering
    \caption{Results under varying sample size \textit{n} and number of variables within nonlinear environments.}
    \label{RMS_varying}
  \resizebox{1\linewidth}{!}{
	\begin{tabular}{c|lcc|lcc|lcc} 
		\toprule
			 & \multicolumn{3}{c|}{n=1000, m=5} & \multicolumn{3}{c|}{n=1000, m=10}   & \multicolumn{3}{c}{n=1000, m=15}   \\ 
   \cmidrule(r){2-10}
		& $\beta_{S}$ Error & $\beta_{V}$ Error       & $\beta$ Error & $\beta_{S}$ Error & $\beta_{V}$ Error & $\beta$ Error & $\beta_{S}$ Error       & $\beta_{V}$ Error & $\beta$ Error \\ 
  \midrule
        OLS & 3.357  & 0.430  & 1.894  & 3.605  & 0.729  & 2.167  & 3.823  & 0.866  & 2.345   \\ 
        Lasso & 3.390  & 0.326  & 1.858  & 3.586  & 0.647  & 2.117  & 3.940  & 0.390  & 2.165   \\ 
        Ridge & 3.357  & 0.430  & 1.893  & 3.604  & 0.729  & 2.166  & 3.822  & 0.866  & 2.344   \\ 
        SVM & 2.067  & 0.240  & 1.153  & 2.273  & 0.375  & 1.324  & 2.366  & 0.410  & 1.388   \\ 
        DWR & 2.279  & 0.249  & 1.264  & 2.566  & 0.658  & 1.612  & 3.258  & 1.182  & 2.220   \\ 
        DWR\_SVM & 1.799  & 0.303  & 1.051  & 2.077  & 0.483  & 1.280  & 2.494  & 0.918  & 1.706   \\ 
        OUR & \textbf{1.555}  & \textbf{0.199}  & \textbf{0.877}  & \textbf{1.898}  & \textbf{0.373}  & \textbf{1.135}  & \textbf{2.265}  & \textbf{0.382}  & \textbf{1.323}   \\ 
  \midrule
         & \multicolumn{3}{c|}{n=2000, m=5} & \multicolumn{3}{c|}{n=2000, m=10}   & \multicolumn{3}{c}{n=2000, m=15} \\
        \cmidrule(r){2-10}
		& $\beta_{S}$ Error & $\beta_{V}$ Error       & $\beta$ Error & $\beta_{S}$ Error & $\beta_{V}$ Error & $\beta$ Error & $\beta_{S}$ Error       & $\beta_{V}$ Error & $\beta$ Error  \\ 
    \midrule

        OLS & 3.253  & 0.444  & 1.849  & 3.521  & 0.630  & 2.075  & 4.071  & 0.561  & 2.316   \\ 
        Lasso & 3.278  & 0.250  & 1.764  & 3.490  & 0.473  & 1.982  & 4.260  & 0.168  & 2.214   \\ 
        Ridge & 3.253  & 0.444  & 1.848  & 3.520  & 0.630  & 2.075  & 4.071  & 0.561  & 2.316   \\ 
        DWR & 2.147  & 0.231  & 1.189  & 2.244  & 0.493  & 1.369  & 2.749  & 0.974  & 1.861   \\ 
        SVM & 2.020  & 0.271  & 1.145  & 2.158  & 0.315  & 1.237  & 2.453  & 0.349  & 1.401   \\ 
        DWR\_SVM & 1.675  & 0.305  & 0.990  & 1.861  & 0.407  & 1.134  & 2.317  & 0.572  & 1.445   \\ 
        OUR & \textbf{1.544}  & \textbf{0.214}  & \textbf{0.879}  & \textbf{1.719}  & \textbf{0.292}  & \textbf{1.006}  & \textbf{2.125}  & \textbf{0.323}  & \textbf{1.224}   \\ 

    \midrule
         & \multicolumn{3}{c|}{n=3000, m=5} & \multicolumn{3}{c|}{n=3000, m=10}   & \multicolumn{3}{c}{n=3000, m=15} \\
        \cmidrule(r){2-10}
		& $\beta_{S}$ Error & $\beta_{V}$ Error       & $\beta$ Error & $\beta_{S}$ Error & $\beta_{V}$ Error & $\beta$ Error & $\beta_{S}$ Error       & $\beta_{V}$ Error & $\beta$ Error  \\ 
    \midrule
        OLS & 3.297  & 0.335  & 1.816  & 3.593  & 0.579  & 2.086  & 3.736  & 0.611  & 2.173   \\ 
        Lasso & 3.279  & 0.074  & 1.677  & 3.803  & 0.179  & 1.991  & 3.703  & 0.527  & 2.115   \\ 
        Ridge & 3.297  & 0.335  & 1.816  & 3.593  & 0.579  & 2.086  & 3.735  & 0.611  & 2.173   \\
        DWR & 2.178  & 0.150  & 1.164  & 1.970  & 0.415  & 1.192  & 2.610  & 0.547  & 1.578   \\ 
        SVM & 2.066  & 0.217  & 1.141  & 2.046  & 0.338  & 1.192  & 2.261  & 0.329  & 1.295   \\ 
        DWR\_SVM & 1.764  & 0.284  & 1.024  & 1.833  & 0.312  & 1.072  & 2.082  & 0.484  & 1.283   \\ 
        OUR & \textbf{1.748}  & \textbf{0.065}  & \textbf{0.907}  & \textbf{1.618}  & \textbf{0.171}  & \textbf{0.894}  & \textbf{2.007}  & \textbf{0.325}  & \textbf{1.166}   \\ 
  \bottomrule
	\end{tabular}

 }
\vspace{-1mm}
\end{table*}

\subsection{Validation on Three Real-World Datasets}
To further validate the effectiveness of our model in real-world scenarios, we perform experiments on three different EHR datasets. All data are prepossessed to ensure no sensitive information is exposed. 

\subsubsection{Datasets and Settings}
\textbf{Heart Disease} is retrieved from the repository of the University of California, Irvine~\citep{asuncion2007uci}. We follow previous work to use 13 of 76 attributes: \emph{Age}, \emph{Sex}, \emph{cp}, \emph{threstbps}, \emph{chol}, \emph{fbs}, \emph{restecg}, \emph{thalach}, \emph{exang}, \emph{oldpeak}, \emph{slope}, \emph{cam} and \emph{thal}. 

\textbf{Esophageal Cancer} consists of data from 261 patients who underwent esophagectomy for esophageal cancer between 2009 and 2018. The collected characteristics include patient demographics, medical and surgical history, clinical tumor staging, adjuvant chemoradiotherapy, esophagectomy procedure type, postoperative pathologic tumor staging, adjuvant chemoradiotherapy, postoperative complications, cancer recurrence, and mortality. 

\textbf{Cauda Equina Syndrome (CES)} is extracted from the Statewide Planning and Research Cooperative System (SPARCS)~\citep{new1984statewide}, a comprehensive database of all payers for all hospitalizations in New York State. 
Based on diagnostic and procedure codes, patients with CES who underwent surgery between 2000 and 2015 were selected. Patient demographics (age, gender, race, comorbidities, and insurance status) and hospital characteristics (measured by hospital bed number quartiles).

\begin{table*}[!t]
    \centering
    \caption{Ablative study. $\gamma$ and $\lambda$ are Lagrange penalty operators.}
    \label{Ablative}
  \resizebox{1\linewidth}{!}{
	\begin{tabular}{c|lcc|lcc|lcc} 
		\toprule
			 & \multicolumn{3}{c|}{$\gamma=600$, $\lambda=0.0001$} & \multicolumn{3}{c|}{$\gamma=600$, $\lambda=0.0005$}   & \multicolumn{3}{c}{$\gamma=600$, $\lambda=0.001$}   \\ 
   \cmidrule(r){2-10}
		& $C=0$ & $C=0.5$       & $C=1$ & $C=0$ & $C=0.5$       & $C=1$& $C=0$ & $C=0.5$       & $C=1$ \\ 
  \midrule
        $\beta_{S}$ Error & 1.956  & 1.919  & 1.996  & 1.769  & \textbf{1.926}  & 2.003  & 1.956  & 2.026  & 2.073   \\ 
        $\beta_{V}$ Error & 0.238  & 0.179  & 0.166  & 0.245  & 0.187  & 0.178  & 0.246  & 0.199  & 0.175   \\ 
        $RMSE$ Error & 4.943  & 4.732  & 4.680  & 4.854  & \textbf{4.726}  & 4.675  & 4.951  & 4.856  & 4.808   \\ 
  \midrule
         & \multicolumn{3}{c|}{$\gamma=800$, $\lambda=0.0001$} & \multicolumn{3}{c|}{$\gamma=800$, $\lambda=0.0005$}   & \multicolumn{3}{c}{$\gamma=800$, $\lambda=0.001$} \\
        \cmidrule(r){2-10}
		& $C=0$ & $C=0.5$   & $C=1$ & $C=0$ & $C=0.5$  & $C=1$& $C=0$ & $C=0.5$       & $C=1$ \\ 
    \midrule

        $\beta_{S}$ Error & 1.954  & 2.022  & 2.070  & 1.784  & 2.025  & 2.068  & 1.960  & 2.019  & 2.009   \\ 
        $\beta_{V}$ Error & 0.240  & 0.197  & 0.172  & 0.234  & 0.195  & 0.176  & 0.245  & 0.195  & 0.174   \\ 
        $RMSE$ Error & 4.945  & 4.859  & 4.825  & 4.849  & 4.860  & 4.793  & 4.961  & 4.858  & 4.674   \\ 

    \midrule
         & \multicolumn{3}{c|}{$\gamma=1000$, $\lambda=0.0001$} & \multicolumn{3}{c|}{$\gamma=1000$, $\lambda=0.0005$}   & \multicolumn{3}{c}{$\gamma=1000$, $\lambda=0.001$} \\
        \cmidrule(r){2-10}
		& $C=0$ & $C=0.5$       & $C=1$ & $C=0$ & $C=0.5$       & $C=1$& $C=0$ & $C=0.5$       & $C=1$ \\ 
    \midrule
        $\beta_{S}$ Error & 1.962  & 2.022  & 2.075  & 1.959  & 1.928  & 2.073  & 1.962  & 2.024  & 2.006   \\ 
        $\beta_{V}$ Error & 0.242  & 0.196  & 0.173  & 0.250  & 0.187  & 0.178  & 0.244  & 0.189  & 0.169   \\ 
        $RMSE$ Error & 4.938  & 4.859  & 4.812  & 4.950  & \textbf{4.726}  & 4.811  & 4.947  & 4.854  & 4.672   \\ 
  \bottomrule
	\end{tabular}

 }
\vspace{-1mm}
\end{table*}

\noindent\textbf{Pre-Processing:} We convert the continuous variables into categorical variables before feeding them to the model. To handle missing data in the datasets, we adopted MICE (Multiple imputations by chained equations) by transforming imputation problems into estimation problems where each variable will be regressed on the other variables. This method provides promising flexibility since every variable can be assigned a suitable distribution~\citep{wulff2017multiple}. Then we apply the SMOTE algorithm~\citep{fernandez2018smote} to address the class imbalance issue in our datasets.

\noindent\textbf{Feature Selection:} Redundant information in EHR datasets may cause noise and irrelevant information during feature extraction. A feature selection method~\citep{guyon2002gene} is adopted. To improve the robustness of the model, we divide the dataset randomly into five groups for cross-validation. Each time we extract one group as the test set to analyze and measure the average performance in the feature selection process. Due to the high complexity of our model, we apply and compare the four baseline models: XGboost, SVM, Logistic Regression, and Random Forest to extract important features in feature selection and input the set of the features with the highest average AUROC scores into our model. 
In the end, we extract 13, 47, and 45 features for \emph{Heart Disease}, \emph{Esophageal Cancer}, and \emph{Cauda Equina Syndrome}, respectively.


\noindent\textbf{Comparison to Baseline Models:} After the feature selection process, we transform continuous features into categorical variables before inputting data into our model and then applying one-hot encoding to convert categorical attributes into numeric, since the association rule mining algorithm in our paper cannot accept continuous features. However, the data set, filtered by feature selection, is directly fed into baseline models since forcing the continuous features to be discretized may lead to worse performance of the model. We assign $20\%$ data into test datasets and compare our model with five baseline models: Logistic Regression, Random Forest, XGboost, SVM and MLP as shown in Appendix~\ref{causal_results_scores}

\subsubsection{Results}
To measure the performance of models, we calculate accuracy, precision, recall and F1 scores. The result is shown in Appendix~\ref{causal_results_scores}, Table~\ref{resultaccuracy}. In addition to calculating the metrics of the traditional models, we input the filtered rules as one-zero matrix $X$ into the baselines rather than the original datasets. In \emph{Heart Disease} and \emph{Esophageal Cancer} datasets, rules do help XGBoost, Random Forest and MLP to improve the performance, while in \emph{Cauda Equina Syndrome} datasets rules can improve the performance of all models. For SVM and Logistic Regression, the effect of the model can be greatly improved after combining the rules. Our model generally performs the best on all three datasets, similar to the SVM performance, while achieving high interpretability as discussed below. 

Combining the experiments in the previous section, better performance is not equivalent to obtaining the real rules. To compare the causality calculated by our model and baselines, we ask three groups of doctors of the corresponding domains to score each rule. Three groups of doctors are from Cardiology, ENT and Neurosurgery departments, and each group consists of three doctors. we apply the models to calculate the importance of features to score each rule. To verify rating consistency between our model and doctors leveraging Spearman Coefficients. Results can be found in Table~\ref{resultaccuracy}. As can be observed, causality rankings of the baseline models vary greatly, indicating unstable performance. In these datasets, the causal value of our model is higher than other baselines, implying that the scoring of our model is more consistent with the standard of doctors.

\begin{table*}[!t]
    \centering
    \caption{Prediction performances over various healthcare datasets.}
    \label{resultaccuracy}
  \resizebox{1\linewidth}{!}{
	\begin{tabular}{clcccc|lccccc} 
		\toprule
			  &\multicolumn{5}{|c|}{Non Rule-based} & \multicolumn{6}{c}{Rule-based} \\ 
   \cmidrule(r){2-12}
      &\multicolumn{1}{|c}{XGBoost} & RF & SVM & LR & MLP & XGBoost & RF & SVM & LR & MLP & Ours  \\ 
     \cmidrule(r){1-12}
      &\multicolumn{11}{c}{Heart Disease} \\
      \cmidrule(r){1-12}
          \multicolumn{1}{c|}{Accuracy} & 0.903 & 0.887 & 0.885 & 0.869 & 0.947 & 0.869 & 0.868 & 0.960 & 0.934 & 0.878 & \textbf{0.960}  \\ 
          \multicolumn{1}{c|}{F1} & 0.88 & 0.863 & 0.899 & 0.882 & 0.952 & 0.882 & 0.879 & 0.963 & 0.94 & 0.892 & \textbf{0.964}  \\ 
          \multicolumn{1}{c|}{Precision} & 0.88 & 0.846 & 0.886 & 0.882 & 0.941 & 0.857 & 0.864 & \textbf{0.972} & 0.939 & 0.85 & 0.966  \\ 
          \multicolumn{1}{c|}{Recall} & 0.88 & 0.88 & 0.912 & 0.882 & 0.963 & 0.909 & 0.897 & 0.956 & 0.945 & 0.94 & \textbf{0.964}  \\ 
          \multicolumn{1}{c|}{Causality} & - & - & - & - & - & 0.398 & 0.274 & 0.455 & 0.458 & 0.402 & \textbf{0.528}  \\ 
         \cmidrule(r){1-12}
      &\multicolumn{11}{c}{Esophageal Cancer} \\
      \cmidrule(r){1-12}
         \multicolumn{1}{c|}{Accuracy} & 0.788 & 0.75 & 0.827 & 0.808 & 0.75 & 0.738 & 0.727 & 0.900 & 0.812 & 0.846 & \textbf{0.900}  \\ 
          \multicolumn{1}{c|}{F1} & 0.776 & 0.683 & 0.809 & 0.8 & 0.735 & 0.708 & 0.697 & \textbf{0.888} & 0.783 & 0.824 & 0.885  \\ 
          \multicolumn{1}{c|}{Precision} & 0.704 & 0.737 & 0.827 & 0.808 & 0.72 & 0.723 & 0.692 & 0.867 & 0.804 & 0.843 & \textbf{0.874}  \\ 
          \multicolumn{1}{c|}{Recall} & 0.864 & 0.636 & 0.792 & 0.833 & 0.75 & 0.699 & 0.713 & \textbf{0.913} & 0.771 & 0.812 & 0.900  \\ 
          \multicolumn{1}{c|}{Causality} & - & - & - & - & - & 0.13 & 0.236 & 0.281 & 0.327 & 0.16 & \textbf{0.339}  \\ 
         \cmidrule(r){1-12}
      &\multicolumn{11}{c}{Cauda Equina Syndrome} \\
      \cmidrule(r){1-12}
         \multicolumn{1}{c|}{Accuracy} & 0.788 & 0.75 & 0.827 & 0.808 & 0.75 & 0.883 & 0.779 & 0.887 & 0.886 & 0.891 & \textbf{0.893}  \\
          \multicolumn{1}{c|}{F1} & 0.776 & 0.683 & 0.809 & 0.8 & 0.735 & 0.88 & 0.78 & 0.883 & 0.882 & 0.888 & \textbf{0.888}  \\ 
          \multicolumn{1}{c|}{Precision} & 0.704 & 0.737 & 0.827 & 0.808 & 0.72 & 0.818 & 0.706 & 0.825 & 0.822 & 0.831 & \textbf{0.834}  \\ 
          \multicolumn{1}{c|}{Recall} & 0.864 & 0.636 & 0.792 & 0.833 & 0.75 & 0.951 & 874 & 0.95 & 0.952 & \textbf{0.953} & 0.951  \\ 
         \multicolumn{1}{c|}{Causality} & - & - & - & - & - &0.231 & 0.298 & 0.279 & 0.132 & 0.262 & \textbf{0.477}\\

  \bottomrule
	\end{tabular}

 }
\vspace{-1mm}
\end{table*}

\section{Conclusion}
In this paper, we present a causal inference approach focusing on interpretability and nonlinear environments for healthcare applications. The proposed method extracts association rules from the raw features as new representations to be used by our model. A novel regularizer that is capable for handling both linear and nonlinear confoundings is constructed to enable our model's adaption to real-world applications. The superior performances on a synthetic dataset and three real-world EHR datasets from different domains compared to baseline methods validate both the effectiveness and generalizability of the proposed method. Consistent ratings with healthcare professionals on the extracted rules further validate the model's interpretability, while not sacrificing accuracy.


\bibliographystyle{iclr2023_conference}
\bibliography{mybibfile}

\begin{thebibliography}{47}
\providecommand{\natexlab}[1]{#1}
\providecommand{\url}[1]{\texttt{#1}}
\expandafter\ifx\csname urlstyle\endcsname\relax
  \providecommand{\doi}[1]{doi: #1}\else
  \providecommand{\doi}{doi: \begingroup \urlstyle{rm}\Url}\fi

\bibitem[Agatonovic-Kustrin \& Beresford(2000)Agatonovic-Kustrin and
  Beresford]{agatonovic2000basic}
S~Agatonovic-Kustrin and Rosemary Beresford.
\newblock Basic concepts of artificial neural network (ann) modeling and its
  application in pharmaceutical research.
\newblock \emph{Journal of pharmaceutical and biomedical analysis}, 22\penalty0
  (5):\penalty0 717--727, 2000.

\bibitem[Agrawal et~al.(1994)Agrawal, Srikant, et~al.]{agrawal1994fast}
Rakesh Agrawal, Ramakrishnan Srikant, et~al.
\newblock Fast algorithms for mining association rules.
\newblock In \emph{Proc. 20th int. conf. very large data bases, VLDB}, volume
  1215, pp.\  487--499. Citeseer, 1994.

\bibitem[Ahmad et~al.(2018)Ahmad, Eckert, and
  Teredesai]{ahmad2018interpretable}
Muhammad~Aurangzeb Ahmad, Carly Eckert, and Ankur Teredesai.
\newblock Interpretable machine learning in healthcare.
\newblock In \emph{Proceedings of the 2018 ACM international conference on
  bioinformatics, computational biology, and health informatics}, pp.\
  559--560, 2018.

\bibitem[Ahmed et~al.(2021)Ahmed, Jeon, and Piccialli]{ahmed2021deep}
Imran Ahmed, Gwanggil Jeon, and Francesco Piccialli.
\newblock A deep-learning-based smart healthcare system for patient’s
  discomfort detection at the edge of internet of things.
\newblock \emph{IEEE Internet of Things Journal}, 8\penalty0 (13):\penalty0
  10318--10326, 2021.

\bibitem[Asuncion \& Newman(2007)Asuncion and Newman]{asuncion2007uci}
Arthur Asuncion and David Newman.
\newblock Uci machine learning repository, 2007.

\bibitem[Bayardo~Jr \& Agrawal(1999)Bayardo~Jr and Agrawal]{bayardo1999mining}
Roberto~J Bayardo~Jr and Rakesh Agrawal.
\newblock Mining the most interesting rules.
\newblock In \emph{Proceedings of the fifth ACM SIGKDD international conference
  on Knowledge discovery and data mining}, pp.\  145--154, 1999.

\bibitem[Bollapragada et~al.(2018)Bollapragada, Nocedal, Mudigere, Shi, and
  Tang]{bollapragada2018progressive}
Raghu Bollapragada, Jorge Nocedal, Dheevatsa Mudigere, Hao-Jun Shi, and Ping
  Tak~Peter Tang.
\newblock A progressive batching l-bfgs method for machine learning.
\newblock In \emph{International Conference on Machine Learning}, pp.\
  620--629. PMLR, 2018.

\bibitem[Borgelt \& Kruse(2002)Borgelt and Kruse]{borgelt2002induction}
Christian Borgelt and Rudolf Kruse.
\newblock Induction of association rules: Apriori implementation.
\newblock In \emph{Compstat}, pp.\  395--400. Springer, 2002.

\bibitem[Chen \& Guestrin(2016)Chen and Guestrin]{chen2016xgboost}
Tianqi Chen and Carlos Guestrin.
\newblock Xgboost: A scalable tree boosting system.
\newblock In \emph{Proceedings of the 22nd acm sigkdd international conference
  on knowledge discovery and data mining}, pp.\  785--794, 2016.

\bibitem[Croskerry(2013)]{croskerry2013mindless}
Pat Croskerry.
\newblock From mindless to mindful practice—cognitive bias and clinical
  decision making.
\newblock \emph{N Engl J Med}, 368\penalty0 (26):\penalty0 2445--2448, 2013.

\bibitem[Cui \& Athey(2022)Cui and Athey]{cui2022stable}
Peng Cui and Susan Athey.
\newblock Stable learning establishes some common ground between causal
  inference and machine learning.
\newblock \emph{Nature Machine Intelligence}, 4\penalty0 (2):\penalty0
  110--115, 2022.

\bibitem[Du et~al.(2019)Du, Liu, and Hu]{du2019techniques}
Mengnan Du, Ninghao Liu, and Xia Hu.
\newblock Techniques for interpretable machine learning.
\newblock \emph{Communications of the ACM}, 63\penalty0 (1):\penalty0 68--77,
  2019.

\bibitem[Fern{\'a}ndez et~al.(2018)Fern{\'a}ndez, Garcia, Herrera, and
  Chawla]{fernandez2018smote}
Alberto Fern{\'a}ndez, Salvador Garcia, Francisco Herrera, and Nitesh~V Chawla.
\newblock Smote for learning from imbalanced data: progress and challenges,
  marking the 15-year anniversary.
\newblock \emph{Journal of artificial intelligence research}, 61:\penalty0
  863--905, 2018.

\bibitem[Gandhi et~al.(2006)Gandhi, Kachalia, Thomas, Puopolo, Yoon, Brennan,
  and Studdert]{gandhi2006missed}
Tejal~K Gandhi, Allen Kachalia, Eric~J Thomas, Ann~Louise Puopolo, Catherine
  Yoon, Troyen~A Brennan, and David~M Studdert.
\newblock Missed and delayed diagnoses in the ambulatory setting: a study of
  closed malpractice claims.
\newblock \emph{Annals of internal medicine}, 145\penalty0 (7):\penalty0
  488--496, 2006.

\bibitem[Guyon et~al.(2002)Guyon, Weston, Barnhill, and Vapnik]{guyon2002gene}
Isabelle Guyon, Jason Weston, Stephen Barnhill, and Vladimir Vapnik.
\newblock Gene selection for cancer classification using support vector
  machines.
\newblock \emph{Machine learning}, 46\penalty0 (1):\penalty0 389--422, 2002.

\bibitem[Han et~al.(2000)Han, Pei, and Yin]{han2000mining}
Jiawei Han, Jian Pei, and Yiwen Yin.
\newblock Mining frequent patterns without candidate generation.
\newblock \emph{ACM sigmod record}, 29\penalty0 (2):\penalty0 1--12, 2000.

\bibitem[Hastie \& Tibshirani(2017)Hastie and
  Tibshirani]{hastie2017generalized}
Trevor~J Hastie and Robert~J Tibshirani.
\newblock \emph{Generalized additive models}.
\newblock Routledge, 2017.

\bibitem[Herpertz et~al.(2017)Herpertz, Huprich, Bohus, Chanen, Goodman,
  Mehlum, Moran, Newton-Howes, Scott, and Sharp]{herpertz2017challenge}
Sabine~C Herpertz, Steven~K Huprich, Martin Bohus, Andrew Chanen, Marianne
  Goodman, Lars Mehlum, Paul Moran, Giles Newton-Howes, Lori Scott, and Carla
  Sharp.
\newblock The challenge of transforming the diagnostic system of personality
  disorders.
\newblock \emph{Journal of personality disorders}, 31\penalty0 (5):\penalty0
  577--589, 2017.

\bibitem[Hoerl \& Kennard(1970)Hoerl and Kennard]{hoerl1970ridge}
Arthur~E Hoerl and Robert~W Kennard.
\newblock Ridge regression: applications to nonorthogonal problems.
\newblock \emph{Technometrics}, 12\penalty0 (1):\penalty0 69--82, 1970.

\bibitem[Hutcheson(2011)]{hutcheson2011ordinary}
Graeme~D Hutcheson.
\newblock Ordinary least-squares regression.
\newblock \emph{L. Moutinho and GD Hutcheson, The SAGE dictionary of
  quantitative management research}, pp.\  224--228, 2011.

\bibitem[Imbens \& Rubin(2015)Imbens and Rubin]{imbens2015causal}
Guido~W Imbens and Donald~B Rubin.
\newblock \emph{Causal inference in statistics, social, and biomedical
  sciences}.
\newblock Cambridge University Press, 2015.

\bibitem[Kuang et~al.(2018)Kuang, Cui, Athey, Xiong, and Li]{kuang2018stable}
Kun Kuang, Peng Cui, Susan Athey, Ruoxuan Xiong, and Bo~Li.
\newblock Stable prediction across unknown environments.
\newblock In \emph{Proceedings of the 24th ACM SIGKDD International Conference
  on Knowledge Discovery \& Data Mining}, pp.\  1617--1626, 2018.

\bibitem[Kuang et~al.(2020{\natexlab{a}})Kuang, Li, Geng, Xu, Zhang, Liao,
  Huang, Ding, Miao, and Jiang]{kuang2020causal}
Kun Kuang, Lian Li, Zhi Geng, Lei Xu, Kun Zhang, Beishui Liao, Huaxin Huang,
  Peng Ding, Wang Miao, and Zhichao Jiang.
\newblock Causal inference.
\newblock \emph{Engineering}, 6\penalty0 (3):\penalty0 253--263,
  2020{\natexlab{a}}.

\bibitem[Kuang et~al.(2020{\natexlab{b}})Kuang, Xiong, Cui, Athey, and
  Li]{kuang2020stable}
Kun Kuang, Ruoxuan Xiong, Peng Cui, Susan Athey, and Bo~Li.
\newblock Stable prediction with model misspecification and agnostic
  distribution shift.
\newblock In \emph{Proceedings of the AAAI Conference on Artificial
  Intelligence}, volume~34, pp.\  4485--4492, 2020{\natexlab{b}}.

\bibitem[Kuang et~al.(2021)Kuang, Zhang, Wu, Wu, Zhuang, and
  Zhang]{kuang2021balance}
Kun Kuang, Hengtao Zhang, Runze Wu, Fei Wu, Yueting Zhuang, and Aijun Zhang.
\newblock Balance-subsampled stable prediction across unknown test data.
\newblock \emph{ACM Transactions on Knowledge Discovery from Data (TKDD)},
  16\penalty0 (3):\penalty0 1--21, 2021.

\bibitem[Lee \& Siau(2001)Lee and Siau]{lee2001review}
Sang~Jun Lee and Keng Siau.
\newblock A review of data mining techniques.
\newblock \emph{Industrial Management \& Data Systems}, 2001.

\bibitem[Lou et~al.(2013)Lou, Caruana, Gehrke, and Hooker]{lou2013accurate}
Yin Lou, Rich Caruana, Johannes Gehrke, and Giles Hooker.
\newblock Accurate intelligible models with pairwise interactions.
\newblock In \emph{Proceedings of the 19th ACM SIGKDD international conference
  on Knowledge discovery and data mining}, pp.\  623--631, 2013.

\bibitem[Ma et~al.(2021)Ma, Dong, Huang, Mietchen, and Li]{ma2021assessing}
Jing Ma, Yushun Dong, Zheng Huang, Daniel Mietchen, and Jundong Li.
\newblock Assessing the causal impact of covid-19 related policies on outbreak
  dynamics: A case study in the us.
\newblock \emph{arXiv preprint arXiv:2106.01315}, 2021.

\bibitem[Muandet et~al.(2013)Muandet, Balduzzi, and
  Sch{\"o}lkopf]{muandet2013domain}
Krikamol Muandet, David Balduzzi, and Bernhard Sch{\"o}lkopf.
\newblock Domain generalization via invariant feature representation.
\newblock In \emph{International Conference on Machine Learning}, pp.\  10--18.
  PMLR, 2013.

\bibitem[of~Health \& Bureau(1984)of~Health and Bureau]{new1984statewide}
New York (State).~Department of~Health and New York (State).~SPARCS Bureau.
\newblock \emph{Statewide Planning and Research Cooperative System Annual
  Report Series}.
\newblock New York State Department of Health, 1984.

\bibitem[Ordonez et~al.(2006)Ordonez, Ezquerra, and
  Santana]{ordonez2006constraining}
Carlos Ordonez, Norberto Ezquerra, and Cesar~A Santana.
\newblock Constraining and summarizing association rules in medical data.
\newblock \emph{Knowledge and information systems}, 9\penalty0 (3):\penalty0
  1--2, 2006.

\bibitem[Pal(2005)]{pal2005random}
Mahesh Pal.
\newblock Random forest classifier for remote sensing classification.
\newblock \emph{International journal of remote sensing}, 26\penalty0
  (1):\penalty0 217--222, 2005.

\bibitem[Pearl(2009)]{pearl2009causal}
Judea Pearl.
\newblock Causal inference in statistics: An overview.
\newblock \emph{Statistics surveys}, 3:\penalty0 96--146, 2009.

\bibitem[Pearl(2018)]{pearl2018theoretical}
Judea Pearl.
\newblock Theoretical impediments to machine learning with seven sparks from
  the causal revolution.
\newblock \emph{arXiv preprint arXiv:1801.04016}, 2018.

\bibitem[Per{\c{c}}{\i}n et~al.(2019)Per{\c{c}}{\i}n, Ya{\u{g}}in,
  G{\"u}ldo{\u{g}}an, and Yolo{\u{g}}lu]{perccin2019arm}
{\.I}brahim Per{\c{c}}{\i}n, Fatma~Hilal Ya{\u{g}}in, Emek G{\"u}ldo{\u{g}}an,
  and Saim Yolo{\u{g}}lu.
\newblock Arm: An interactive web software for association rules mining and an
  application in medicine.
\newblock In \emph{2019 International Artificial Intelligence and Data
  Processing Symposium (IDAP)}, pp.\  1--5. IEEE, 2019.

\bibitem[Rojas-Carulla et~al.(2018)Rojas-Carulla, Sch{\"o}lkopf, Turner, and
  Peters]{rojas2018invariant}
Mateo Rojas-Carulla, Bernhard Sch{\"o}lkopf, Richard Turner, and Jonas Peters.
\newblock Invariant models for causal transfer learning.
\newblock \emph{The Journal of Machine Learning Research}, 19\penalty0
  (1):\penalty0 1309--1342, 2018.

\bibitem[Royce et~al.(2019)Royce, Hayes, and Schwartzstein]{royce2019teaching}
Celeste~S Royce, Margaret~M Hayes, and Richard~M Schwartzstein.
\newblock Teaching critical thinking: a case for instruction in cognitive
  biases to reduce diagnostic errors and improve patient safety.
\newblock \emph{Academic Medicine}, 94\penalty0 (2):\penalty0 187--194, 2019.

\bibitem[Rudin(2019)]{rudin2019stop}
Cynthia Rudin.
\newblock Stop explaining black box machine learning models for high stakes
  decisions and use interpretable models instead.
\newblock \emph{Nature Machine Intelligence}, 1\penalty0 (5):\penalty0
  206--215, 2019.

\bibitem[Shen et~al.(2020)Shen, Cui, Liu, Zhang, Li, and Chen]{shen2020stable}
Zheyan Shen, Peng Cui, Jiashuo Liu, Tong Zhang, Bo~Li, and Zhitang Chen.
\newblock Stable learning via differentiated variable decorrelation.
\newblock In \emph{Proceedings of the 26th ACM SIGKDD International Conference
  on Knowledge Discovery \& Data Mining}, pp.\  2185--2193, 2020.

\bibitem[Sornalakshmi et~al.(2021)Sornalakshmi, Balamurali, Venkatesulu,
  Krishnan, Ramasamy, Kadry, and Lim]{sornalakshmi2021efficient}
M~Sornalakshmi, S~Balamurali, M~Venkatesulu, M~Navaneetha Krishnan,
  Lakshmana~Kumar Ramasamy, Seifedine Kadry, and Sangsoon Lim.
\newblock An efficient apriori algorithm for frequent pattern mining using
  mapreduce in healthcare data.
\newblock \emph{Bulletin of Electrical Engineering and Informatics},
  10\penalty0 (1):\penalty0 390--403, 2021.

\bibitem[Suykens \& Vandewalle(1999)Suykens and Vandewalle]{suykens1999least}
Johan~AK Suykens and Joos Vandewalle.
\newblock Least squares support vector machine classifiers.
\newblock \emph{Neural processing letters}, 9\penalty0 (3):\penalty0 293--300,
  1999.

\bibitem[Tibshirani(1996)]{tibshirani1996regression}
Robert Tibshirani.
\newblock Regression shrinkage and selection via the lasso.
\newblock \emph{Journal of the Royal Statistical Society: Series B
  (Methodological)}, 58\penalty0 (1):\penalty0 267--288, 1996.

\bibitem[Wulff \& Jeppesen(2017)Wulff and Jeppesen]{wulff2017multiple}
Jesper~N Wulff and Linda~Ejlskov Jeppesen.
\newblock Multiple imputation by chained equations in praxis: guidelines and
  review.
\newblock \emph{Electronic Journal of Business Research Methods}, 15\penalty0
  (1):\penalty0 41--56, 2017.

\bibitem[Xu et~al.(2021)Xu, Cui, Shen, Zhang, and Zhang]{xu2021stable}
Renzhe Xu, Peng Cui, Zheyan Shen, Xingxuan Zhang, and Tong Zhang.
\newblock Why stable learning works? a theory of covariate shift
  generalization.
\newblock \emph{arXiv preprint arXiv:2111.02355}, 2021.

\bibitem[Yuan(2017)]{yuan2017improved}
Xiuli Yuan.
\newblock An improved apriori algorithm for mining association rules.
\newblock In \emph{AIP conference proceedings}, volume 1820, pp.\  080005. AIP
  Publishing LLC, 2017.

\bibitem[Zafar \& Khan(2019)Zafar and Khan]{zafar2019dlime}
Muhammad~Rehman Zafar and Naimul~Mefraz Khan.
\newblock Dlime: A deterministic local interpretable model-agnostic
  explanations approach for computer-aided diagnosis systems.
\newblock \emph{arXiv preprint arXiv:1906.10263}, 2019.

\bibitem[Zhang et~al.(2021)Zhang, Cui, Xu, Zhou, He, and Shen]{zhang2021deep}
Xingxuan Zhang, Peng Cui, Renzhe Xu, Linjun Zhou, Yue He, and Zheyan Shen.
\newblock Deep stable learning for out-of-distribution generalization.
\newblock In \emph{Proceedings of the IEEE/CVF Conference on Computer Vision
  and Pattern Recognition}, pp.\  5372--5382, 2021.

\end{thebibliography}

\newpage
\appendix
\section{Appendix}
\subsection{Algorithm}\label{app_algorithm}

We combine algorithm~\ref{rulereItemduce} with object function~\ref{selection} to select the robust rules and prune the redundant items. In the \emph{RulesSelection} function, we delete one rule each time with lowest $\|w\|^{2}_{2}$ and save the rule sets with the highest accuracy. In the \emph{ItemReduce} function, we apply cross-validation to train SVM model and save the item sets with best accuracy.
\renewcommand{\algorithmicrequire}{\textbf{Input:}}  
\renewcommand{\algorithmicensure}{\textbf{Output:}} 
\begin{algorithm}
        \caption{Rules Selection and Item Reduction}
        \label{rulereItemduce}
        \begin{algorithmic}[1] 
            
            \Require $Rules\{X_{i}\}$ are the association rules obtained by Apriori algorithm with training datasets. $data$ is EHR datasets.
            \Ensure $Bestrules$
            \Function{RulesSelection}{$Rules, data$}
                \State $Bestrules \gets Rules$
                \State $Objfunction$ is objective function
                \State $Select \gets Bestrules$
                \State $Bestaccuracy \gets Select$
                \State $Lastrules \gets \varnothing$
                \While{$Select \neq Lastrules$}
                    \State $Lastrules \gets Select$
                    \State $w \gets \operatorname*{argmin} Objectfunction(Select, data)$
                    \State $Selected \gets \operatorname*{argmin} w_{i}^{2}$
                    \State $Temprules \gets \{Bestrules\}/\{Selected\}$
                    \State $Tempaccuracy \gets Temprules$
                    \If{$Tempaccuracy > Bestaccuracy$}
                        \State $Bestaccuracy \gets Tempaccuracy$
                        \State $Select \gets Temprules$
                    \EndIf
                \EndWhile
                \State \Return{$Bestrules$}
            \EndFunction
            \Function{ItemReduce}{$Bestrules, data$}
                \State $Bestauc \gets SVM(Bestrules, data)$
                \State $Lastrules \gets \varnothing$
                \While{$Bestrules \neq Lastrules$}
                    \State $Item \gets \operatorname*{argmax} SVM(\{Bestrules\}/\{Item\})$
                    \State $Accuracy \gets SVM(\{Bestrules\}/\{Item\})$
                    \If{$Accuracy \geq Bestauc$}
                        \State $Bestauc \gets Accuracy$
                        \State $Bestrules \gets \{Bestrules\}/\{Item\}$
                    \EndIf
                \EndWhile
                \State \Return{$Bestrules$}
            \EndFunction
        \end{algorithmic}
\end{algorithm}

\newpage
\subsection{Datasets and Settings}\label{exp_datasets}
In addition to the linear settings, we propose to include nonlinear evaluations under a nonlinear environment:

\textbf{Linear Environment:} \quad For this setting, we construct features $\mathbf{S}$ that causes unstable $\mathbf{V}$ by auxiliary variables $z$ with linear relationship among features only.
\begin{equation*}
\begin{array}{r}
\mathbf{Z}_{, 1}, \cdots, \mathbf{Z}_{, p} \stackrel{i i d}{\sim} \mathcal{N}(0,1), \mathbf{X}_{, 1}, \cdots, \mathbf{X}_{, p_{v}} \stackrel{i i d}{\sim} \mathcal{N}(0,1) \\
\mathbf{S}_{, i}=0.8 * \mathbf{Z}_{, i}+0.2 * \mathbf{Z}_{, i+1}, i=1,2, \cdots, p_{s}
\end{array}
\end{equation*}
\begin{equation*}
\mathbf{V}_{\cdot, j}=0.8 * \mathbf{X}_{\cdot, j}+0.2 * \mathbf{X}_{\cdot, j+1}+\mathcal{N}(0,1)
\end{equation*}

\textbf{Nonlinear Environment:} \quad In this setting, we combined square relationship and exponential relationship to generate various environment including potential nonlinear confounding to test our reweighted regularizer.
\begin{equation*}
\begin{aligned}
\mathbf{V}_{\cdot, j}&=\mathbf{X}_{\cdot, j} + 0.4 * \mathbf{X}_{\cdot, j+1} + 0.4 * exp(\mathbf{X}_{\cdot, j+1}) \\
&+ 0.4 * \mathbf{X}^{2}_{\cdot, j+1}+ 0.1 * \mathbf{X}^{3}_{\cdot, j+1} + \mathcal{N}(0,1) \\
\mathbf{S}_{\cdot, j} &=\mathbf{Z}_{\cdot, j}+0.4 * \mathbf{Z}_{\cdot, j+1} + 0.4 * exp(\mathbf{Z}_{\cdot, j+1})\\
&+ 0.4 * \mathbf{Z}^{2}_{\cdot, j+1} + 0.1 * \mathbf{Z}^{3}_{\cdot, j+1} + \mathcal{N}(0,1)
\end{aligned}
\end{equation*}

To further test the robustness of our algorithm, we assume that there are unobserved nonlinear terms, and construct the label $Y$ as shown in Equation~\ref{experiment}. Combined with weighed SVM loss function, we train our model to estimate the regression coefficient $\beta$. In this experiment, we set $\beta_{s}=\left\{\frac{1}{3},-\frac{2}{3}, 1,-\frac{1}{3}, \frac{2}{3},-1, \cdots\right\}, \beta_{v}=\overrightarrow{0}$, and $\varepsilon=$ $\mathcal{N}(0,0.3)$.
\begin{equation}
\label{experiment}
Y_{p o l y}=f(\mathbf{S})+\varepsilon=[\mathbf{S}, \mathbf{V}] \cdot\left[\beta_{s}, \beta_{v}\right]^{T}+\mathbf{S}_{\cdot, 1} \mathbf{S}_{\cdot, 2} + \varepsilon
\end{equation}

\noindent\textbf{Baselines:} \quad We compare our model with five baseline methods:

\begin{itemize}
    \item Ordinary Least Square (OLS)~\citep{hutcheson2011ordinary}:
    \begin{equation*}
        \min \|Y-\mathbf{X} \beta\|_2^2
    \end{equation*}
    \item Lasso~\citep{tibshirani1996regression}:
    \begin{equation*}
        \min \|Y-\mathrm{X} \beta\|_2^2+\lambda_1\|\beta\|_1
    \end{equation*}
    \item Ridge~\citep{hoerl1970ridge}:
    \begin{equation*}
        \min \|Y-\mathrm{X} \beta\|_2^2+\lambda_1\|\beta\|_2
    \end{equation*}
    \item Decorrelated Weighting Regression (DWR)~\citep{kuang2020stable}:
    \begin{equation*}
    \begin{array}{ll} 
        & \min _{W, \beta} \sum_{i=1}^n W_i \cdot\left(Y_i-\mathbf{X}_{i,} \beta\right)^2 \\
        \text { s.t } & \sum_{j=1}^p\left\|\mathbf{X}_{, j}^T \boldsymbol{\Sigma}_W \mathbf{X}_{,-j} / n-\mathbf{X}_{, j}^T W / n \cdot \mathbf{X}_{,-j}^T W / n\right\|_2^2<\lambda_2 \\
    \end{array}
    \end{equation*}
    \item Support Vector Machines (SVM)~\citep{suykens1999least}:
        \begin{equation*}
            \min _{w, b, \zeta, \zeta^*} \frac{1}{2} w^T w+ \sum_{i=1}^n \left(\zeta_i+\zeta_i^*\right)
        \end{equation*}
    \item SVM combined with DWR(DWR\_SVM):
    \begin{equation*}
    \begin{array}{ll}
        &\min _{w, b, \zeta, \zeta^*} \frac{1}{2} w^T w+ \sum_{i=1}^n W_{i}\left(\zeta_i+\zeta_i^*\right) \\
        &\text { s.t } \sum_{j=1}^p\left\|\mathbf{X}_{, j}^T \boldsymbol{\Sigma}_W \mathbf{X}_{,-j} / n-\mathbf{X}_{, j}^T W / n \cdot \mathbf{X}_{,-j}^T W / n\right\|_2^2<\lambda_2 \\
    \end{array}
    \end{equation*}
\end{itemize}

\begin{figure*}
	\includegraphics[width=1\textwidth]{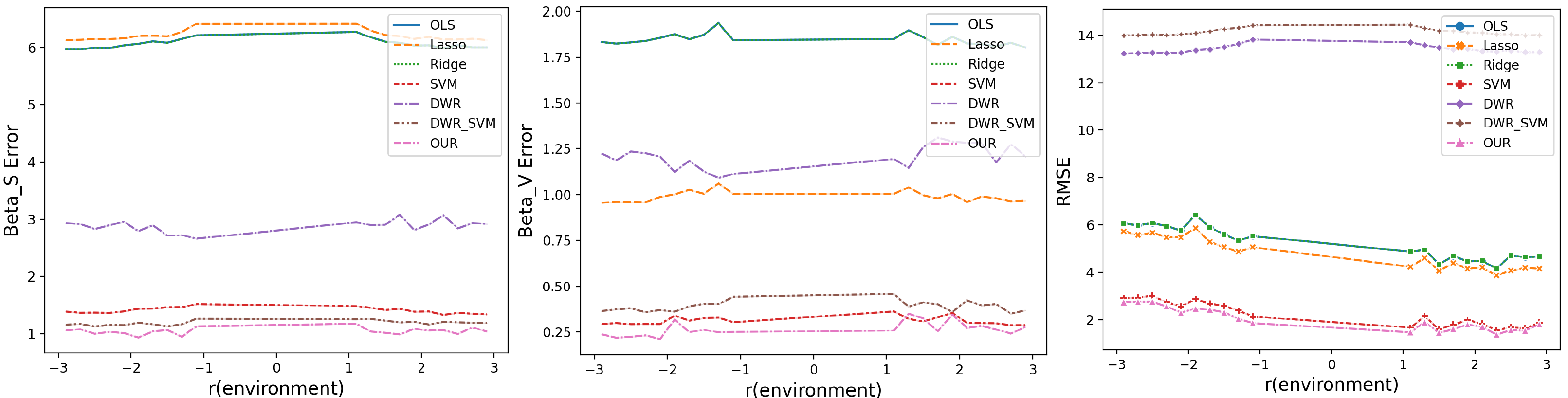}
	\centering
	\caption{Figures describe the $\beta_{S}$, $\beta_{V}$ and RMSE with various environments.}
	\label{shift}
\end{figure*}

\noindent\textbf{Generating Various Environments} To test the stability of the algorithms, we generate a set of environment $e$ with a distinct distribution distribution $P_{XY}$. Following the Kuang's experiment~\cite{kuang2020stable}, we generate different environments based on various $P(S|V)$. To simplify the problem, we simulate $P(S_{b}|V$ on a subset $S_{b} \in S$, where the dimension of $S_{b}$ is $0.2*p$. We applied the bias rate equation $P r=\prod_{\mathbf{S}_i \in \mathbf{S}_b}|r|^{-5 * D_{i}}$ to tune the $P(S_{b}|V$, where $D_i=\left|f(\mathbf{S})-\operatorname{sign}(r) * \mathbf{V}_i\right|, r \in[-3,-1) \cup(1,3]$. $r > 1$ indicates that $Y$ and $S_{b}$ have positive unstable relationships, while $r < -1$ corresponds to the negative unstable relationships. The higher absolute value of $r$ the stronger connection between $S_{b}$ and $Y$, leading to generate different environments. The result is shown in Figure~\ref{shift}.

\newpage
\subsection{Causal Experiment Results and Explanation}\label{causal_results_scores}

We compare our model with five traditional methods:

\begin{itemize}
    \item \textbf{Logistic Regression} \quad We leverage the logistic regression classifier with L-BFGS solver for classification~\citep{bollapragada2018progressive}.
    \item \textbf{Random Forest} \quad We apply standard Random Forest classifier to solve the classification problem~\citep{pal2005random}.
    \item \textbf{XGboost} \quad We adopt XGBoost, an extreme gradient boosting methods, to compare with other models~\citep{chen2016xgboost}.
    \item \textbf{SVM} \quad We apply supervised learning models, SVM, with linear kernel to analyze data for classification~\citep{suykens1999least}.
    \item \textbf{MLP} \quad We use the traditional neural network multi-layer perceptron to solve this classification task~\citep{agatonovic2000basic}.
\end{itemize}
We sort the rules in descending order by calculating the importance and show the top five rules compared with the doctor's score~\ref{toprules}. The details of the description for each feature in the rules are shown in the Table~\ref{description}. The scoring criteria are as follows:
\begin{itemize}
    \item \textbf{Score 4:} \quad Strongly agree that the rule contains causality.
    \item \textbf{Score 3:} \quad Agree that the rule contains causality.
    \item \textbf{Score 2:} \quad Disagree with this rule.
    \item \textbf{Score 1:} \quad Strongly disagree with this rule.
\end{itemize}

\begin{table}[!htp]
\centering
\caption{Rules filtered by algorithm are sorted in a descending order by our algorithm compared with the scores given by doctors.}
\label{toprules}
\resizebox{1\linewidth}{!}{
\begin{tabular}{llllllc}
\toprule
\multicolumn{6}{l}{Association Rules}  & Scores  \\ \midrule
\multicolumn{6}{l}{\textbf{Heart Disease}} & \\
\multicolumn{6}{l}{age middle, \#major vessels0, fixed defect, pressure normal, ST-T wave abnormality $\Rightarrow$ heart disease}  &  4  \\
 \multicolumn{6}{l}{age middle, cholesterol edge, \#major vessels0, lower than 120mg/ml $\Rightarrow$ heart disease}  & 3  \\
\multicolumn{6}{l}{non-anginal pain, cholesterol high, no exercise induced angina $\Rightarrow$ heart disease}  & 4  \\
\multicolumn{6}{l}{ST-T wave abnormality, downsloping $\Rightarrow$ heart disease}  & 4  \\
 \multicolumn{6}{l}{fixed defect, \#major vessels0, cholesterol edge $\Rightarrow$ heart disease}  & 4  \\
 \midrule
 \multicolumn{6}{l}{\textbf{Esophageal Cancer}} & \\
 \multicolumn{6}{l}{Modified Ryan Score 2.0, Esophagectomy Procedure 4 $\Rightarrow$ recurrence} & 2  \\ 
 \multicolumn{6}{l}{tobacco use, Alcohol Use, Neoadjuvant Radiation, Histological Grade 2, Final Histology 1  $\Rightarrow$ recurrence} & 4  \\ 
\multicolumn{6}{l}{Histological Grade 3, Neoadjuvant Radiation, Esophagectomy Procedure 4, Final Histology 1 $\Rightarrow$ recurrence}  & 4  \\ 
\multicolumn{6}{l}{clinical m Stage 1, Histological Grade 3, Neoadjuvant Radiation, Esophagectomy Procedure 4, Final Histology 1 $\Rightarrow$ recurrence}  & 4  \\ 
\multicolumn{6}{l}{esoph tumor location 4, Esophagectomy Procedure 5, Histological Grade 3 $\Rightarrow$ recurrence}   & 3  \\
 \midrule
\multicolumn{6}{l}{\textbf{Cauda Equina Syndrome}} & \\
\multicolumn{6}{l}{elixsum, beds, procedure 03 09 $\Rightarrow$ die360}  & 4  \\ 
\multicolumn{6}{l}{Emergency, diagnosis 344 60, complication 240days $\Rightarrow$ die360}  & 4  \\ 
\multicolumn{6}{l}{diagnosis 344 60, life threatening, complication 240days $\Rightarrow$ die360}  & 4  \\ 
\multicolumn{6}{l}{if aa $\Rightarrow$ die360} & 4  \\ 
\multicolumn{6}{l}{or potentially disabling conditions, complication 240days $\Rightarrow$ die360}  & 4  \\ 
\midrule
\end{tabular}
}
\vspace{-1mm}
\end{table}

\newpage
\subsection{Introduction for Features}
\begin{table}[!htp]
\centering
\caption{Introduction of individual features on different datasets.}
\label{description}
\resizebox{1\linewidth}{!}{
\begin{tabular}{llllllllllll}
\toprule
\multicolumn{3}{l}{Features}  & \multicolumn{9}{l}{Explanation}  \\ \midrule
\multicolumn{3}{l}{\textbf{Heart Disease}} &  \omit\\
\multicolumn{3}{l}{\textbf{age middle}} & \multicolumn{9}{l}{Patients between the ages of 40 and 60} \\
\multicolumn{3}{l}{\textbf{\#major vessels0}} & \multicolumn{9}{l}{The number of major vessels (0-3) colored by flourosopy is 0} \\
\multicolumn{3}{l}{\textbf{fixed defect}} & \multicolumn{9}{l}{Thalium stress test result is fixed defect} \\
\multicolumn{3}{l}{\textbf{pressure normal}} & \multicolumn{9}{l}{Blood pressure within the normal range} \\
\multicolumn{3}{l}{\textbf{ST-T wave abnormality}} & \multicolumn{9}{l}{Resting electrocardiography result is ST-T wave abnormality} \\
\multicolumn{3}{l}{\textbf{cholesterol edge}} & \multicolumn{9}{l}{Serum cholesterol is in range $(200, 220]$ mg/dl} \\
 \multicolumn{3}{l}{\textbf{lower than 120mg/ml}} & \multicolumn{9}{l}{Fasting blood sugar is lower than 120mg/ml} \\
 \multicolumn{3}{l}{\textbf{non-anginal pain}} & \multicolumn{9}{l}{Chest pain type is  non-angina} \\
 \multicolumn{3}{l}{\textbf{cholesterol high}} & \multicolumn{9}{l}{Serum cholesterol is higher than 220 mg$/$dl} \\
 \multicolumn{3}{l}{\textbf{no exercise induced angina}} & \multicolumn{9}{l}{not Exercise induced angina} \\
  \multicolumn{3}{l}{\textbf{downsloping}} & \multicolumn{9}{l}{Slope of peak exercise ST segment is downsloping} \\
  \multicolumn{3}{l}{\textbf{heart disease}} & \multicolumn{9}{l}{It refers to the presence of heart disease in the patient} \\
 \midrule
 \multicolumn{3}{l}{\textbf{Esophageal Cancer}} & \omit\\
\multicolumn{3}{l}{\textbf{Modified Ryan Score 2.0}} & \multicolumn{9}{l}{(near complete response): single cells or rare small groups of cancer cells} \\
\multicolumn{3}{l}{\textbf{Esophagectomy Procedure 4}} & \multicolumn{9}{l}{Complete MIS/Robotic McKeown (Three-Hole) esophagectomy} \\
\multicolumn{3}{l}{\textbf{tobacco use}} & \multicolumn{9}{l}{Use tobacco} \\
\multicolumn{3}{l}{\textbf{Alcohol Use}} & \multicolumn{9}{l}{Use Alcohol} \\
\multicolumn{3}{l}{\textbf{Neoadjuvant Radiation}} & \multicolumn{9}{l}{Patient underwent neoadjuvant radiation} \\
\multicolumn{3}{l}{\textbf{Histological Grade 2}} & \multicolumn{9}{l}{How differentiated the tumor is: Moderately Differentiated} \\
\multicolumn{3}{l}{\textbf{Final Histology 1}} & \multicolumn{9}{l}{History: Adenocarcinoma} \\
\multicolumn{3}{l}{\textbf{Histological Grade 3}} & \multicolumn{9}{l}{How differentiated the tumor is: Poorly Differentiated} \\
\multicolumn{3}{l}{\textbf{clinical m Stage 1}} & \multicolumn{9}{l}{Details any spread (metastasis) to other sites of the body: M0} \\
 \multicolumn{3}{l}{\textbf{esoph tumor location 4}} & \multicolumn{9}{l}{Lower Thoracic, including GE junction} \\
\multicolumn{3}{l}{\textbf{Esophagectomy Procedure 5}} & \multicolumn{9}{l}{Hybrid (Laparoscopy $+$ Thoracotomy) McKeown (Three-Hole) esophagectomy} \\
\multicolumn{3}{l}{\textbf{recurrence}} & \multicolumn{9}{l}{Details whether the patient experience recurrence of their cancer} \\
\midrule
\multicolumn{3}{l}{\textbf{Cauda Equina Syndrome}} & \omit\\
\multicolumn{3}{l}{\textbf{elixsum}} & \multicolumn{9}{l}{Elixhauser comorbidity sum for that patient is high} \\
\multicolumn{3}{l}{\textbf{beds}} & \multicolumn{9}{l}{Number of beds in the hospital is small} \\
\multicolumn{3}{l}{\textbf{procedure 03 09}} & \multicolumn{9}{l}{ICD-9-CM Procedure Codes: 03.09} \\
\multicolumn{3}{l}{\textbf{Emergency}} & \multicolumn{9}{l}{The patient requires immediate medical intervention as a result of severe} \\
\multicolumn{3}{l}{\textbf{diagnosis 344 60}} & \multicolumn{9}{l}{ICD9 indicators} \\
\multicolumn{3}{l}{\textbf{complication 240days}} & \multicolumn{9}{l}{Indicators for complication within 240 days of discharge} \\
\multicolumn{3}{l}{\textbf{life threatening}} & \multicolumn{9}{l}{The patient's condition is very dangerous} \\
\multicolumn{3}{l}{\textbf{if aa}} & \multicolumn{9}{l}{The racial of the patient is {\tt African American}} \\
\multicolumn{3}{l}{\textbf{die360}} &  \multicolumn{9}{l}{Patient died within 360 days} \\
\midrule
\end{tabular}
}
\end{table}

\end{document}